\documentclass{article} % For LaTeX2e
\usepackage[final]{colm2026_conference}
\usepackage{tcolorbox}
\usepackage{xurl}
\usepackage{microtype}
\usepackage{hyperref}
\usepackage{booktabs}
\usepackage{multirow}
\usepackage{array}
\usepackage{listings}
\usepackage{xcolor}
\usepackage{graphicx}
\usepackage[table]{xcolor}
\lstdefinestyle{prompt}{
    basicstyle=\ttfamily\scriptsize,%
     breaklines=true,%
     breakindent=0pt,%
     columns=fullflexible,%
     keepspaces=true,%
     showstringspaces=false,%
     frame=single,%
     framesep=6pt,%
     xleftmargin=6pt,%
     xrightmargin=6pt,%
   }
\usepackage[T1]{fontenc}

\usepackage{lineno}

\definecolor{darkblue}{rgb}{0, 0, 0.5}
\definecolor{darkred}{HTML}{914A46}
\hypersetup{colorlinks=true, citecolor=darkblue, linkcolor=darkblue, urlcolor=darkblue}

\title{LLMs and the Homogenization of Teaching\thanks{An earlier version of this paper appeared at COLM 2026 under the title ``Can LLMs Match Human Diversity in Educational Content Generation?''}}

\author{Mei Tan\thanks{Co-first authors}, Hyunji Nam\footnotemark[2], Hari Subramonyam \& Dorottya Demszky\\
Stanford University\\
Stanford, CA 94305, USA \\
\texttt{\{mxtan,hjnam,harihars,ddemszky\}@stanford.edu}
}

\begin{document}

\ifcolmsubmission
\linenumbers
\fi

\maketitle

\begin{abstract}
Large language models (LLMs) are known to lack diversity in open-ended text generation, a limitation that is especially consequential in education, where homogeneity can narrow the kinds of thinking and disciplinary work students are invited to do. We study diversity in two educational generation tasks: math items and essay feedback. We compare frontier and open-weight models with human-authored reference distributions using lexical, semantic, and domain-grounded measures. For math, we examine the mathematical activities students are asked to perform; for feedback, we examine the revision work requested and the rhetorical strategies used to communicate it. We evaluate direct prompting, two diversity prompting methods, and post-training. Across both tasks, directly prompted model outputs are markedly less diverse than human-authored content---78\% higher on lexical (Self-BLEU) and 51\% higher on semantic (embedding) similarity---and represent a narrower range of instructional forms. Diversity prompting and post-training increase variation largely through surface features without broadening the instructional repertoire. Models inherit and amplify the most normative routines of human instruction. Overcoming mode collapse in education will require richer instructional data and investment in teachers’ creative capacity and professional judgment.
\end{abstract}

\section{Introduction}

Educational materials shape the kinds of thinking students are invited to do~\citep{marton1997learning, marton2014necessary, kullberg2017made}. A math problem can ask students to compute a value, construct a representation, evaluate a claim, or analyze a strategy~\citep{stein1998mathematical, swan2005improving, burkhardt2013task}. A feedback comment can ask a writer to correct an error, elaborate an idea, or evaluate the effectiveness of a rhetorical choice~\citep{bereiter1987psychology, straub1997students, carless2018development}. Diversity in educational content captures the breadth of learning opportunities and disciplinary work that materials make available. 

\begin{figure*}[h]
    \centering
    \includegraphics[width=\textwidth]{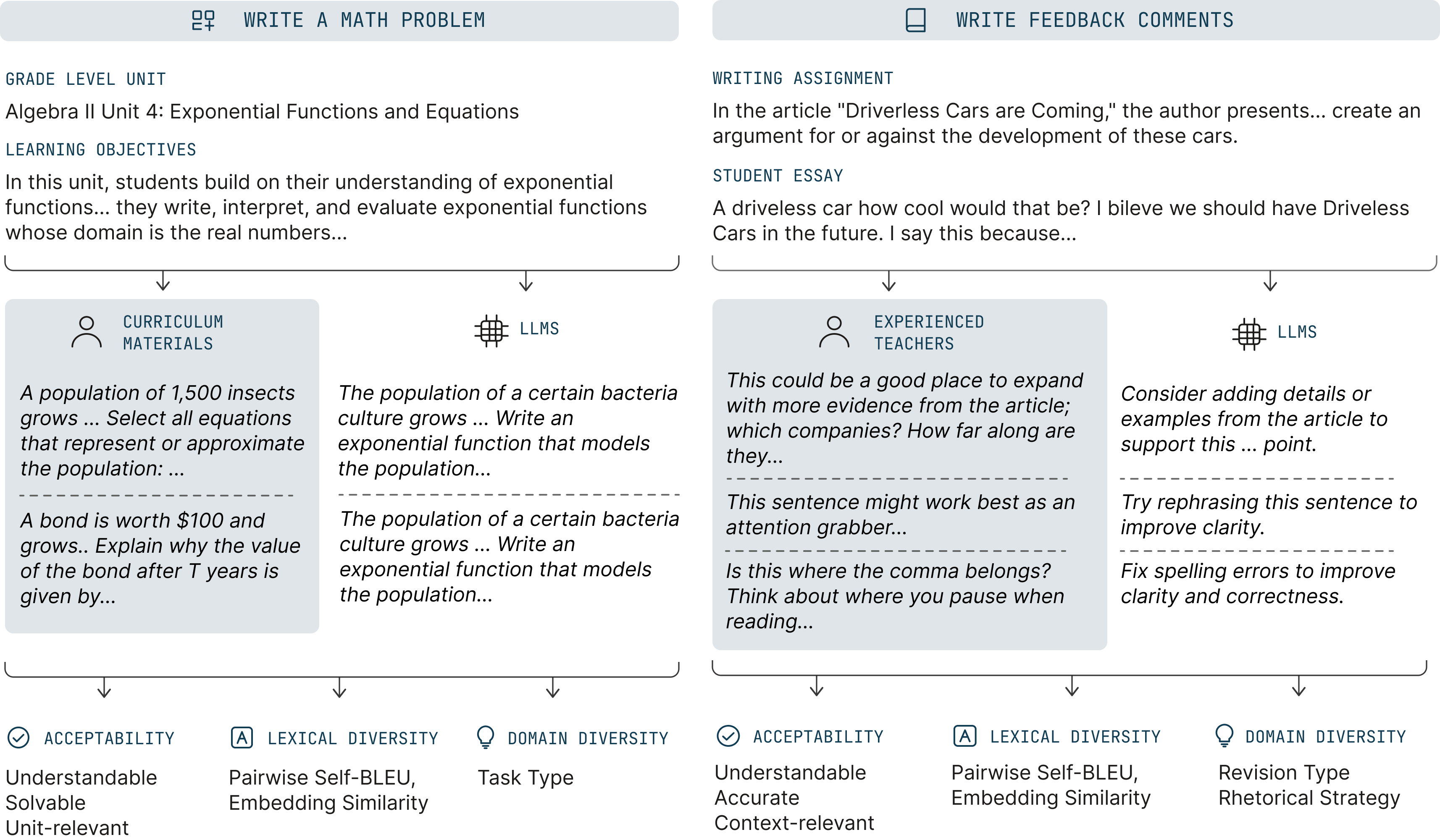}
    \caption{We study diversity in two educational generation tasks: math item generation and essay feedback.}
    \label{fig:process}
    \vspace{-0.3cm}
\end{figure*}

As large language models are increasingly used to generate educational materials, including tutoring responses, lesson plans, assessments, and feedback~\citep{kasneci2023chatgpt, ai4ed_2024, team2024learnlm, khan2023harnessing}, recent work has raised concerns about their performance in these instructional roles. Studies have documented outputs that are inaccurate, poorly aligned with curricular standards, inappropriate for target learners, or reproduce harmful stereotypes~\citep{steiss2024comparing, bulut2024automatic, scaria2024automated, lucy2024mathfishevaluatinglanguagemodel, tan2026markedpedagogies}. Yet even when generated content meets basic standards of correctness and relevance, homogenization can limit its instructional value. A system that draws from a narrow repertoire of pedagogical moves may appear competent while repeatedly offering the same kinds of tasks, feedback, and invitations for student thinking~\citep{stein2009implementing}.

This concern connects to a broader literature on LLM output homogeneity and mode collapse. Across domains, models have been shown to reproduce a narrow subset of the human response distribution~\citep{jiang2025artificialhivemindopenendedhomogeneity, wu2024generativemonoculture, shumailov2024ai, xu2025echoesinai, namuduri2025qudsimquantifyingdiscoursesimilarities}. This collapse may reflect a typicality bias in post-training~\citep{li2024entropic, liu2026alignmenttax, kirk2023understanding} as well as the modal nature of human demonstrations, shaped by repetition of familiar and institutional routines~\citep{lortie1975schoolteacher, bourdieu1977outline}.

We study this problem in two popular educational generation tasks: \textbf{(1) math problems} and \textbf{(2) essay feedback}. We compare outputs from frontier and open-weight models to human-authored reference distributions, examine the effects of diversity prompting and post-training, and evaluate diversity using lexical, semantic, and domain-grounded measures. We assess the acceptability of outputs and examine how their language constructs instructional priorities and invitations for student thinking.

Across tasks, models fail to recover the breadth of instructional forms present in human-authored content. but largely by finding new words for already overrepresented routines, such as asking students to make corrections in writing or perform computations in math. This further concentrates materials in a narrow set of instructional practices, crowding out less common but more ambitious forms of teaching that ask students to connect, synthesize, and evaluate ideas.

Our contributions are:
\begin{itemize}
\item We frame diversity in educational generation in terms of the breadth of learning opportunities and disciplinary work made available to students.
\item We study diversity in two educational generation tasks using human expert-authored content as reference distributions.
\item We combine lexical, semantic, and domain-grounded measures to distinguish surface variation from pedagogically meaningful variation.
\item We show that models fail to recover the breadth observed in human-authored reference distributions, which are themselves constrained.
\item We show that diversity prompting and post-training can improve diversity scores while further concentrating outputs in already overrepresented instructional practices.
\end{itemize}

\section{Related Works}
\textbf{LLM output homogeneity.} A growing body of work has documented homogeneity in LLM-generated text~\citep{jiang2025artificialhivemindopenendedhomogeneity, wu2024generativemonoculture, shumailov2024ai}. \citet{jiang2025artificialhivemindopenendedhomogeneity} show that 70+ LLMs exhibit high pairwise embedding similarity within and across model families, \citet{wu2024generativemonoculture} find sentiment collapse in book reviews, and \citet{xu2025echoesinai} find repetitive plot elements in stories. Similar patterns appear in narrative discourse~\citep{namuduri2025qudsimquantifyingdiscoursesimilarities}, scientific ideation~\citep{ruan2026evaluating}, and code generation~\citep{lu-etal-2025-benchmarking}. 

Post-training methods, including supervised fine-tuning (SFT), reinforcement learning from human feedback (RLHF), and direct preference optimization (DPO), have been shown to reduce output diversity~\citep{li2024entropic, kirk2023understanding, liu2026alignmenttax}. Recent work has modified post-training to reward more varied responses. DivPO selects rare responses as chosen examples and common responses as rejected examples~\citep{lanchantin2025divpo}, reporting improved diversity in persona generation and creative writing. Other work proposes inference-time methods~\citep{khan2026seeing}. Verbalized sampling asks models to generate a distribution of possible responses along with their probabilities~\citep{zhang2025verbalizedsamplingmitigatemode} and negatively correlated sampling prompts models to generate candidate responses that differ from one another~\citep{zhang2026cultivatingpluralismalgorithmicmonoculture}.

\textbf{Measuring LLM output diversity.} Lexical diversity is commonly measured using Self-BLEU~\citep{zhu2018texygenbenchmarkingplatformtext} and distinct-$n$~\citep{li2016diversity}. Both metrics rely on n-gram overlap and can reward superficial rephrasing of repeated content. Embedding-based approaches, such as BERTScore~\citep{zhang2020bertscore}, the Vendi Score~\citep{friedman2023vendiscore}, and MAUVE~\citep{pillutla2023mauve}, use learned representations to compare the semantic content of texts. However, neither lexical nor embedding-based metrics capture whether diversity is consequential in practice~\citep{tevet2021evaluating}. This limitation has motivated task-dependent approaches to diversity evaluation. For example, \citet{jain2025taskdependent} define ``functional diversity'' in terms of whether users perceive responses as meaningfully different in practice.

\textbf{LLMs for education.} Research on LLMs in education has explored their use for tutoring, lesson planning, assessment, and feedback, highlighting the potential for more immediate and scalable instructional support~\citep{kasneci2023chatgpt, ai4ed_2024, team2024learnlm, khan2023harnessing}. Evaluations of generated educational content have identified problems with accuracy, curricular alignment, prioritization of issues, and bias~\citep{bulut2024automatic, scaria2024automated, lucy2024mathfishevaluatinglanguagemodel, weissburg2025biasedteachers, du2025benchmarking, steiss2024comparing, tan2026markedpedagogies}. Across educational tasks, studies report generated content that is generic, underspecified, or weakly tailored to the instructional context~\citep{ lammert2024better, hu2025exploring, liu2025discerning, yoon2023evaluation}. We examine this generic-ness as a form of output homogeneity, in which models repeatedly draw on a narrow range of instructional moves.

\section{Data}
\subsection{Human Demonstrations}
\paragraph{Math items.} 
We use a subset of the public MathFish~\citep{lucy2024mathfishevaluatinglanguagemodel} dataset, consisting of standalone, end-of-lesson practice problems from Illustrative Mathematics, a high-quality, open-source K--12 curriculum. We removed duplicate items, excluded problems that depended on images or tables, and converted mathematical expressions from MathJax to ASCII notation for readability. We selected the 20 units with the largest number of eligible items, spanning Grade 6 through High School Algebra II. The resulting dataset contains 1,477 unique math problems.

\paragraph{Essay feedback.}
We use an expanded version of a dataset introduced in prior work \citep{tan2024reframing}. The dataset comprises 3,021 inline excerpt-comment feedback written by 61 experienced U.S. middle and high school ELA teachers. Teachers had an average of 8.25 years of teaching experience and were recruited through professional development programs focused on writing instruction. The comments address 448 student essays responding to 18 writing assignments drawn from the publicly available PERSUADE \citep{persuade_dataset} and ASAP-AES \citep{asap-aes} datasets.

\subsection{LLM Demonstrations}
We evaluate frontier and open-weight models across multiple scales: GPT-5.2, Gemini-3.5-Flash, Llama-3.3-70B Versatile, Qwen-3.6-27B, Llama-3.1-8B-Instruct, and Qwen-2.5-7B-Instruct. We generate LLM outputs under three settings: direct prompting, prompting methods for increasing diversity, and post-training. We use temperature $\tau=1$ with model-default top-$p$. We used OpenAI's inference API for GPT-5.2, Gemini's API for Gemini-3.5-Flash, and Groq for serving Llama-3.3-70B Versatile and Qwen 3.6-27B.

\paragraph{Direct prompting.}
We use direct prompting as a baseline generation condition. In each task, the models receive contextual information associated with the corresponding human demonstrations. For essay feedback, models are given a student essay and its associated writing assignment and asked to generate inline feedback comments. For math items, models are given the Illustrative Mathematics description for the target unit, which summarizes the relevant learning objectives, and asked to generate practice problems. 

\paragraph{Diversity prompting.}
We compare direct prompting with two prompting methods designed to reduce output homogeneity. \textbf{Verbalized sampling} asks the model to generate $N$ possible responses together with an estimated probability for each response~\citep{zhang2025verbalizedsamplingmitigatemode}. \textbf{Negatively correlated sampling} asks the model to generate $N$ responses jointly, with each response differing from the others in the set~\citep{zhang2026cultivatingpluralismalgorithmicmonoculture}. Figures~\ref{fig:direct-prompt} and \ref{fig:sampling-prompt} show the prompts used for each condition.

% -------------------- Figure: direct prompting ----------------------
\begin{figure}[t]
\centering
\begin{lstlisting}[style=prompt]
=== Math Items ===
You are a K-12 math curriculum specialist. Write one practice problem aligned to the given mathematical learning purpose. The problem should be situated within this lesson and contain only one primary mathematical activity. It may address a single aspect of the learning purpose rather than covering the entire lesson. Do not create multiple subproblems or a sequence of distinct tasks that build on one another.
+ [Unit Learning Objective Summary]

=== Essay Feedback ===
You are a middle school English Language Arts teacher. Write inline feedback comments for the given writing assignment prompt and student essay to help your student revise. Each comment must include an exact excerpt from the student's essay and a feedback comment addressing a single topic of revision related to the excerpt.
+ [Writing Assignment Instructions]
+ [Student Essay Text]
\end{lstlisting}
\caption{Direct-prompting system prompts for the math and feedback generation tasks. Prompts abridged for readability. We additionally specified the required output format: math generations contained only the problem text, and feedback generations were returned as structured JSON.}
\label{fig:direct-prompt}
\end{figure}

% -------------------- Figure: sampling variants ----------------------
\begin{figure}[t]
\centering
\begin{lstlisting}[style=prompt]
=== Verbalized Sampling ===
Generate 20 possible responses to the input prompt. Each response must include the response string and the estimated probability 0.0 to 1.0 of this response given the input prompt (relative to the full distribution). Randomly sample the responses from the full distribution.
+ [Direct Prompting User Prompt]

=== Negatively Correlated Sampling ===
Generate 20 possible responses to the input prompt that represent diverse instructional values.
+ [Direct Prompting User Prompt]
\end{lstlisting}
\caption{Sampling-variant prompts. Both wrap the direct-prompting user prompt of Figure~\ref{fig:direct-prompt}.}
\label{fig:sampling-prompt}
\end{figure}

\paragraph{Post-training.}
We evaluate supervised fine-tuning (SFT) and direct preference optimization (DPO)~\citep{rafailov2023directpreferenceoptimizationlanguage} on open-weight models. We use all math items from 10 of the 20 math units and all feedback comments from 9 of the 18 writing assignments for training and holding out the remaining units and assignments for evaluation. For SFT, each training example pairs the task context with a human-authored response. For DPO, we pair a human-authored chosen response with a model-generated rejected response for the same context. We train Qwen-2.5-7B-Instruct and Llama-3.1-8B-Instruct using OpenRLHF~\citep{hu2024openrlhf}. Appendix~\ref{appendix:posttrain} reports the training details and hyperparameters.

\section{Methods}

\subsection{Masking}
Diversity metrics can be inflated by superficial differences, such as changes to variable names in math problems or quotations from student writing in feedback comments. We therefore create a masked version of each output and compute diversity on both the original and masked text. We use GPT-5-nano to identify maskable spans, then audit likely errors through programmatic checks and targeted human review. This procedure identified very few masking errors. Appendix~\ref{appendix:masking} describes the masking prompts, audit procedure, and corrections.

\paragraph{Math items.}
We mask names and labels used in the problem, such as for people, places, points, lines, and shapes, with \texttt{[text variable]}. We mask equation variables, such as $x$, $y$, and $\alpha$, with \texttt{[equation variable]}. We also mask each numeric value with a label indicating its type: \texttt{[whole number]}, \texttt{[decimal number]}, or \texttt{[fraction]}.

\paragraph{Essay feedback.}
We mask direct quotes from the student's essay with \texttt{[quote]} and proposed replacement wording for text in the excerpt with \texttt{[rewrite]}. We also mask new text suggested for the student to add to their work as \texttt{[insertion]}. Although we cannot fully remove essay-specific language, since feedback is inherently reactive to student writing, masking accounts for exact-wording engagement with the essay text. Table~\ref{tab:mask} shows examples.

\begin{table}[htbp]
\centering
\small
\setlength{\tabcolsep}{6pt}
\renewcommand{\arraystretch}{1.15}
\begin{tabular}{@{}
                >{\raggedright\arraybackslash}p{0.08\linewidth}
                >{\raggedright\arraybackslash}p{0.44\linewidth}
                >{\raggedright\arraybackslash}p{0.40\linewidth}
                @{}}
\toprule
\textbf{Domain} & \textbf{Original Output} & \textbf{Masked Output} \\
\midrule
Math
& Jada has a sleeping bag that is rated for 30 degrees F. Write an
  inequality that represents the outdoor temperature at which Jada will
  be able to stay warm\ldots
& [text\_variable] has a sleeping bag that is rated for [whole\_number]
  degrees F. This means that if the temperature outside is at least
  [whole\_number] degrees F, [text\_variable] will\ldots \\
\midrule
Essay
& Check spelling: `assitance' should be `assistance.'
& Check spelling: [quote] should be [rewrite] \\
\midrule
Essay
& Clarify this idea by saying something like, `One good thing about
  driverless cars is that they could help elderly people.'
& Clarify this idea by saying something like, [insertion] \\
\bottomrule
\end{tabular}
\caption{Examples of unmasked and masked outputs for math items and essay
feedback.}
\label{tab:mask}
\end{table}

\subsection{Acceptability Checks}
Because low-quality outputs may still appear diverse, we restrict our diversity analyses to outputs that meet a baseline standard of quality. We apply the same criteria to human-authored and model-generated responses and evaluate acceptability using a panel of three LLM judges: GPT-5-nano, Claude-Haiku-4.5, and Gemini-2.5-Flash-Lite. We manually adjudicate all cases without unanimous agreement. Appendix~\ref{appendix:acceptability} describes these procedures, prompts, common sources of disagreement, and the number of outputs removed.

Across both tasks, outputs must be \textit{understandable}: written in clear language that is complete enough to interpret in context.

\paragraph{Math items.}
We evaluate whether each problem is \textit{solvable} and \textit{unit-relevant}. A problem is solvable if it provides enough information to determine a valid solution. A problem is unit-relevant if it addresses the mathematical content described for the target unit. We manually reviewed all 1,616 cases of judge disagreement, overturning the majority judgment in 26.2\% of cases. Disagreements often involved multiple-choice problems without a correct answer, references to missing diagrams, or valid problems with distractor information or edge-case answers, such as an undefined slope.

\paragraph{Essay feedback.}
We evaluate whether each comment is \textit{accurate} and \textit{context-relevant}. A comment is accurate if its assessment of the student’s writing is true and does not invent or misdiagnose an issue. A comment is context-relevant if it offers a meaningful observation or suggestion to support the student's writing or development as a writer. We used Claude-Sonnet-5 to adjudicate 37,056 cases of judge disagreement, informed by a manual audit of 200 randomly sampled cases. Disagreements often involved feedback that overlooked information elsewhere in the essay, legitimate praise, or direct edits such as single-word corrections. We manually reviewed 5,000 cases in which the adjudicator overturned the majority judgment, agreeing with its decisions in 96\% of cases.

\section{Evaluation Metrics}
\subsection{Self-BLEU}
We measure lexical similarity using mean pairwise BLEU, or Self-BLEU~\citep{zhu2018texygenbenchmarkingplatformtext}. We compute BLEU-3 using NLTK's \texttt{sentence\_bleu} with uniform n-gram weights, method 1 smoothing, and \texttt{wordpunct\_tokenize} without lowercasing. For each instructional unit, we draw 20 random samples of model outputs, each matched to the number of human examples, and average Self-BLEU across samples. We then average across units or assignments. We report results for both original and masked text. 

\subsection{Embedding Similarity}
We measure semantic similarity as the mean pairwise cosine similarity among output embeddings~\citep{jiang2025artificialhivemindopenendedhomogeneity}. We embed each output using \texttt{text-embedding-3-small} and compute similarity within each instructional unit. As with Self-BLEU, we draw 20 model output samples equal in size to the human-authored set for the same unit, average across samples, and then average across units or assignments. We report results for both original and masked text.

\subsection{Domain Measures} We evaluate diversity as coverage over meaningful forms of disciplinary work. We use Claude-Sonnet-5 as an LLM-judge to label the generated data according to codebooks adapted from educational research. In a stratified random sample of 100 outputs per task, manual annotations agreed with Claude's labels in 88\% of Math Task Types, 96\% of Feedback Revision Types, and 92\% of Feedback Rhetorical Strategies. Appendix~\ref{appendix:domainmodels} presents the full codebooks, LLM-judge prompts, and validation procedures.

\begin{table}[htbp]
\centering
\small
\setlength{\tabcolsep}{6pt}
\renewcommand{\arraystretch}{1.15}
\begin{tabular}{@{}
                >{\raggedright\arraybackslash}p{0.13\linewidth}
                >{\raggedright\arraybackslash}p{0.12\linewidth}
                >{\raggedright\arraybackslash}p{0.69\linewidth}
                @{}}
\toprule
\textbf{Measure} & \textbf{Category} & \textbf{Example} \\
\midrule
\multirow{4}{=}{Math Task Type}
& Compute  & Solve \(3x+5=20\) for \(x\). \\
& Construct & Make a dot plot for the given data. \\
& Interpret & If the 5 represents the number of people, what does the 4 represent? \\
& Analyze  & Jana says \(4(x+3)=4x+3\). Explain the mistake. \\
& Connect  & Match each equation to the graph that represents it. \\
& Evaluate & Are \(2b+b\) and \(3b\) equivalent for all values of \(b\)? \\
\midrule
\multirow{4}{=}{Feedback Revision Type}
& Operate     & Fix the spelling of ``necessary.'' \\
& Elaborate   & Consider adding more detail here. \\
& Interpret   & What can we infer about her feelings from this quote? \\
& Connect     & This seems to contradict your earlier point. How can both be true? \\
& Evaluate & Is this the right tone for your audience? \\
\midrule
\multirow{4}{=}{Feedback Rhetorical Strategy}
& Correction       & Change ``their'' to ``there.'' \\
& Criticism        & This is unclear. \\
& Interpretation   & I think you're arguing here that legalization would reduce crime. \\
& Reader Response  & I'm getting lost here on who is speaking! \\
\bottomrule
\end{tabular}
\caption{Selected examples of domain-grounded measures that illustrate the breadth of learning opportunities and disciplinary work that educational materials make available. Additional categories and all definitions and examples are provided in Appendix~\ref{appendix:domainmodels}.}
\label{tab:categories}
\end{table}

% We evaluate diversity as coverage over meaningful forms of disciplinary work. 
% For each task, we define a set of instructional dimensions and use multi-label RoBERTa classifiers to estimate the presence of each dimension in every output. We average these scores across outputs to compare the relative prevalence of each dimension in human-authored and model-generated content. Appendix~\ref{appendix:domainmodels} presents the full codebooks, annotation procedure, and classifier performance.

\paragraph{Math Items.}
We assess \texttt{Task Type}, which captures the mathematical activity students are asked to perform. These activities shape opportunities to engage in different forms of mathematical thinking~\citep{swan2005improving, burkhardt2013task}, such as classifying mathematical objects, interpreting or constructing representations, evaluating statements, creating problems or mathematical objects, and analyzing reasoning. Each problem is assigned a single task type.

\paragraph{Essay Feedback.}
We assess \texttt{Revision Type}, which captures the kind of revision work made available to the student. Comments may ask students to fix an issue, elaborate an idea, synthesize ideas across the essay, or evaluate the effectiveness of their writing~\citep{bereiter1987psychology}. \texttt{Rhetorical Strategy} captures how the feedback is delivered, including open questions, criticism, praise, reader responses, and direct corrections~\citep{straub1997students}. Each comment is assigned all applicable revision types and rhetorical strategies.

\section{Results}

\subsection{LLMs produce more homogeneous content than humans.}
\label{section:1}
Table~\ref{tab:lexical_diversity} compares lexical and semantic diversity in model-generated and human-authored math problems and essay feedback. Across both tasks, all six directly prompted models produce substantially more homogeneous outputs than human references. For math problems, model Self-BLEU is 73–136\% higher than human references, and embedding similarity is 41–112\% higher. For essay feedback, Self-BLEU is 45–98\% higher and embedding similarity is 27–50\% higher.

This diversity gap persists after masking superficial variation. In essay feedback, masking widens the embedding similarity gap for all models, from 27–50\% above human references to 48–59\%, suggesting that models derive more of their apparent diversity from superficial variation than human authors do. 

\begin{table*}[t]
\centering
\small
\setlength{\tabcolsep}{4pt}
\begin{tabular}{@{}l|cccc|cccc@{}}
& \multicolumn{4}{c|}{\bf Math Items} & \multicolumn{4}{c}{\bf Essay Feedback} \\
& \multicolumn{2}{c}{\bf Self-BLEU $\downarrow$} & \multicolumn{2}{c|}{\bf Emb. Sim. $\downarrow$} & \multicolumn{2}{c}{\bf Self-BLEU $\downarrow$} & \multicolumn{2}{c}{\bf Emb. Sim. $\downarrow$} \\
\textbf{Generator} & \textbf{Unmask} & \textbf{Mask} & \textbf{Unmask} & \textbf{Mask} & \textbf{Unmask} & \textbf{Mask} & \textbf{Unmask} & \textbf{Mask} \\
\midrule
\textbf{Human} & 0.394 & 0.602 & 0.347 & 0.479 & 0.270 & 0.318 & 0.246 & 0.257 \\
\midrule
\textbf{GPT-5.2} & \cellcolor{darkred!47}0.852 & \cellcolor{darkred!45}0.879 & \cellcolor{darkred!43}0.651 & \cellcolor{darkred!41}0.693 & \cellcolor{darkred!21}0.460 & \cellcolor{darkred!22}0.527 & \cellcolor{darkred!52}0.369 & \cellcolor{darkred!52}0.409 \\
\quad + Verbalized sampling & \cellcolor{darkred!12}0.516 & \cellcolor{darkred!19}0.716 & \cellcolor{darkred!6}0.391 & \cellcolor{darkred!9}0.528 & \cellcolor{darkred!10}0.358 & \cellcolor{darkred!12}0.433 & \cellcolor{darkred!30}0.316 & \cellcolor{darkred!32}0.351 \\
\quad + Negative sampling & \cellcolor{darkred!17}0.555 & \cellcolor{darkred!15}0.693 & \cellcolor{darkred!8}0.403 & \cellcolor{darkred!10}0.532 & \cellcolor{darkred!10}0.358 & \cellcolor{darkred!12}0.430 & \cellcolor{darkred!30}0.317 & \cellcolor{darkred!32}0.350 \\
\textbf{Gemini 3.5-Flash} & \cellcolor{darkred!42}0.803 & \cellcolor{darkred!40}0.849 & \cellcolor{darkred!27}0.536 & \cellcolor{darkred!24}0.603 & \cellcolor{darkred!24}0.482 & \cellcolor{darkred!24}0.545 & \cellcolor{darkred!28}0.312 & \cellcolor{darkred!42}0.380 \\
\quad + Verbalized sampling & \cellcolor{darkred!7}0.462 & \cellcolor{darkred!2}0.615 & 0.342 & 0.475 & \cellcolor{darkred!15}0.405 & \cellcolor{darkred!17}0.484 & \cellcolor{darkred!21}0.295 & \cellcolor{darkred!44}0.386 \\
\quad + Negative sampling & \cellcolor{darkred!8}0.471 & 0.600 & 0.342 & 0.464 & \cellcolor{darkred!6}0.324 & \cellcolor{darkred!8}0.395 & \cellcolor{darkred!17}0.287 & \cellcolor{darkred!25}0.331 \\
\textbf{Llama-3.3-70B Versatile} & \cellcolor{darkred!50}0.882 & \cellcolor{darkred!49}0.903 & \cellcolor{darkred!45}0.664 & \cellcolor{darkred!47}0.726 & \cellcolor{darkred!29}0.535 & \cellcolor{darkred!32}0.622 & \cellcolor{darkred!41}0.343 & \cellcolor{darkred!45}0.389 \\
\quad + Verbalized sampling & \cellcolor{darkred!39}0.776 & \cellcolor{darkred!45}0.876 & \cellcolor{darkred!8}0.405 & \cellcolor{darkred!14}0.554 & \cellcolor{darkred!33}0.570 & \cellcolor{darkred!32}0.622 & \cellcolor{darkred!42}0.344 & \cellcolor{darkred!36}0.362 \\
\quad + Negative sampling & \cellcolor{darkred!39}0.776 & \cellcolor{darkred!46}0.882 & \cellcolor{darkred!9}0.412 & \cellcolor{darkred!18}0.575 & \cellcolor{darkred!31}0.552 & \cellcolor{darkred!31}0.620 & \cellcolor{darkred!41}0.342 & \cellcolor{darkred!36}0.363 \\
\textbf{Qwen-3.6-27B} & \cellcolor{darkred!30}0.683 & \cellcolor{darkred!25}0.754 & \cellcolor{darkred!20}0.491 & \cellcolor{darkred!16}0.562 & \cellcolor{darkred!13}0.391 & \cellcolor{darkred!27}0.575 & \cellcolor{darkred!30}0.317 & \cellcolor{darkred!51}0.405 \\
\quad + Verbalized sampling & \cellcolor{darkred!9}0.479 & \cellcolor{darkred!12}0.676 & \cellcolor{darkred!2}0.364 & \cellcolor{darkred!7}0.515 & \cellcolor{darkred!4}0.304 & \cellcolor{darkred!23}0.537 & \cellcolor{darkred!17}0.285 & \cellcolor{darkred!37}0.366 \\
\quad + Negative sampling & \cellcolor{darkred!12}0.508 & \cellcolor{darkred!9}0.656 & \cellcolor{darkred!1}0.357 & \cellcolor{darkred!3}0.494 & \cellcolor{darkred!6}0.324 & \cellcolor{darkred!25}0.564 & \cellcolor{darkred!20}0.294 & \cellcolor{darkred!45}0.389 \\
\textbf{Llama-3.1-8B-Instruct} & \cellcolor{darkred!39}0.771 & \cellcolor{darkred!36}0.825 & \cellcolor{darkred!26}0.528 & \cellcolor{darkred!32}0.647 & \cellcolor{darkred!26}0.504 & \cellcolor{darkred!28}0.591 & \cellcolor{darkred!52}0.368 & \cellcolor{darkred!52}0.408 \\
\quad + Verbalized sampling & \cellcolor{darkred!49}0.870 & \cellcolor{darkred!55}0.937 & \cellcolor{darkred!12}0.430 & \cellcolor{darkred!21}0.587 & \cellcolor{darkred!42}0.650 & \cellcolor{darkred!40}0.700 & \cellcolor{darkred!55}0.375 & \cellcolor{darkred!48}0.397 \\
\quad + Negative sampling & \cellcolor{darkred!46}0.842 & \cellcolor{darkred!52}0.923 & \cellcolor{darkred!23}0.512 & \cellcolor{darkred!34}0.657 & \cellcolor{darkred!37}0.607 & \cellcolor{darkred!36}0.663 & \cellcolor{darkred!52}0.367 & \cellcolor{darkred!47}0.393 \\
\quad + SFT & \cellcolor{darkred!29}0.680 & \cellcolor{darkred!30}0.784 & \cellcolor{darkred!11}0.426 & \cellcolor{darkred!17}0.567 & \cellcolor{darkred!31}0.549 & \cellcolor{darkred!24}0.553 & \cellcolor{darkred!18}0.288 & \cellcolor{darkred!46}0.392 \\
\quad + DPO & \cellcolor{darkred!10}0.489 & \cellcolor{darkred!2}0.615 & \cellcolor{darkred!16}0.458 & \cellcolor{darkred!17}0.567 & \cellcolor{darkred!45}0.673 & \cellcolor{darkred!48}0.784 & \cellcolor{darkred!41}0.341 & \cellcolor{darkred!55}0.417 \\
\textbf{Qwen-2.5-7B-Instruct} & \cellcolor{darkred!55}0.931 & \cellcolor{darkred!55}0.939 & \cellcolor{darkred!55}0.735 & \cellcolor{darkred!55}0.765 & \cellcolor{darkred!26}0.503 & \cellcolor{darkred!35}0.652 & \cellcolor{darkred!28}0.312 & \cellcolor{darkred!46}0.390 \\
\quad + Verbalized sampling & \cellcolor{darkred!38}0.764 & \cellcolor{darkred!53}0.925 & \cellcolor{darkred!11}0.422 & \cellcolor{darkred!22}0.594 & \cellcolor{darkred!55}0.766 & \cellcolor{darkred!55}0.848 & \cellcolor{darkred!31}0.319 & \cellcolor{darkred!51}0.406 \\
\quad + Negative sampling & \cellcolor{darkred!40}0.784 & \cellcolor{darkred!49}0.904 & \cellcolor{darkred!23}0.507 & \cellcolor{darkred!42}0.698 & \cellcolor{darkred!51}0.734 & \cellcolor{darkred!50}0.803 & \cellcolor{darkred!30}0.317 & \cellcolor{darkred!40}0.372 \\
\quad + SFT & \cellcolor{darkred!48}0.862 & \cellcolor{darkred!47}0.892 & \cellcolor{darkred!32}0.569 & \cellcolor{darkred!32}0.645 & \cellcolor{darkred!18}0.433 & \cellcolor{darkred!12}0.438 & \cellcolor{darkred!1}0.249 & \cellcolor{darkred!7}0.278 \\
\quad + DPO & \cellcolor{darkred!37}0.760 & \cellcolor{darkred!34}0.810 & \cellcolor{darkred!26}0.529 & \cellcolor{darkred!25}0.608 & \cellcolor{darkred!25}0.494 & \cellcolor{darkred!32}0.627 & \cellcolor{darkred!25}0.304 & \cellcolor{darkred!38}0.367 \\
\bottomrule
\end{tabular}
\caption{\textbf{Lexical and semantic diversity across math items and essay feedback.} We report average pairwise Self-BLEU and embedding similarity (Emb. Sim.) within topic; lower values indicate greater diversity. Masked variants are computed after pre-processing to remove superficial variation.}
\label{tab:lexical_diversity}
\end{table*}

\subsection{LLMs collapse around dominant instructional forms.}
Across both tasks, models amplify the instructional forms already most common in human-authored content, while less common forms become rarer still.

\begin{figure*}[h]
    \centering
    \includegraphics[width=\textwidth]{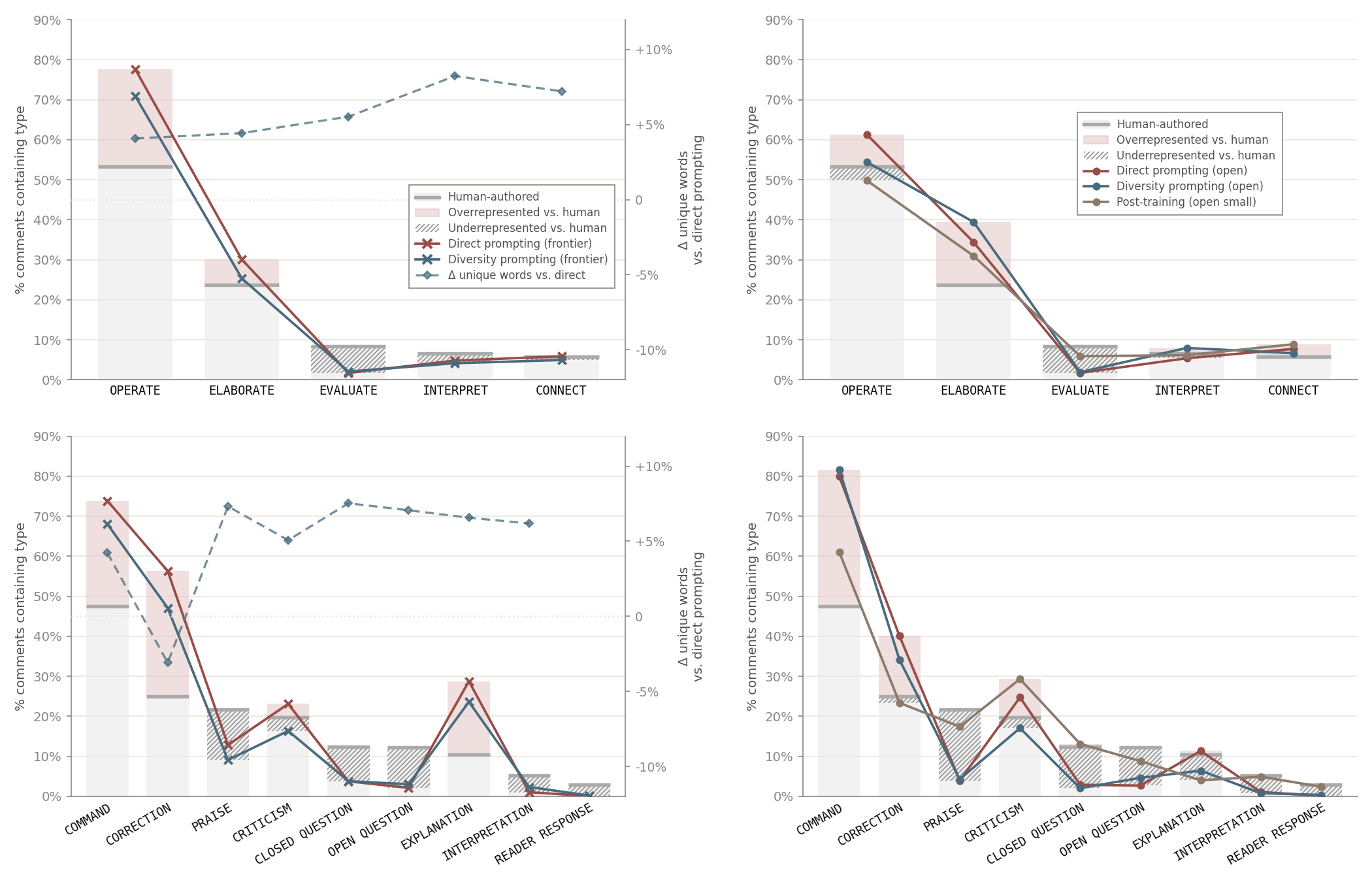}
    \caption{\textbf{Distribution of essay feedback types.} Revision types (top) and rhetorical strategies (bottom) for human-authored and model-generated feedback under direct prompting, diversity prompting, and post-training. Left: frontier models; right: open-weight models. Categories are not mutually exclusive. Dashed lines show the percentage change in unique words relative to direct prompting (right axis).}
    \label{fig:feedbackdomain}
    \vspace{-0.3cm}
\end{figure*}

In essay feedback (Figure~\ref{fig:feedbackdomain}), comments asking students to \textit{operate}, or perform mechanical fixes without judgment or new reasoning, appear in 48\% of human comments but 73\% of model comments under direct prompting. Comments asking students to \textit{elaborate}, or simply produce more text, appear in 18\% of human comments versus 28\% of model comments. Meanwhile, feedback asking students to interpret the implications of their ideas, evaluate their own writing, or connect ideas across the essay is underrepresented. The same collapse occurs for rhetorical strategies, where \textit{corrections} comprise 25\% of human comments but 56\% of model comments. \textit{Commands} occur in 47\% of human comments but are further overrepresented in model comments at 73\%. Open and closed questions and reader responses, already uncommon in human feedback, appear even less frequently in model-generated comments.

\begin{figure*}[h]
    \centering
    \includegraphics[width=\textwidth]{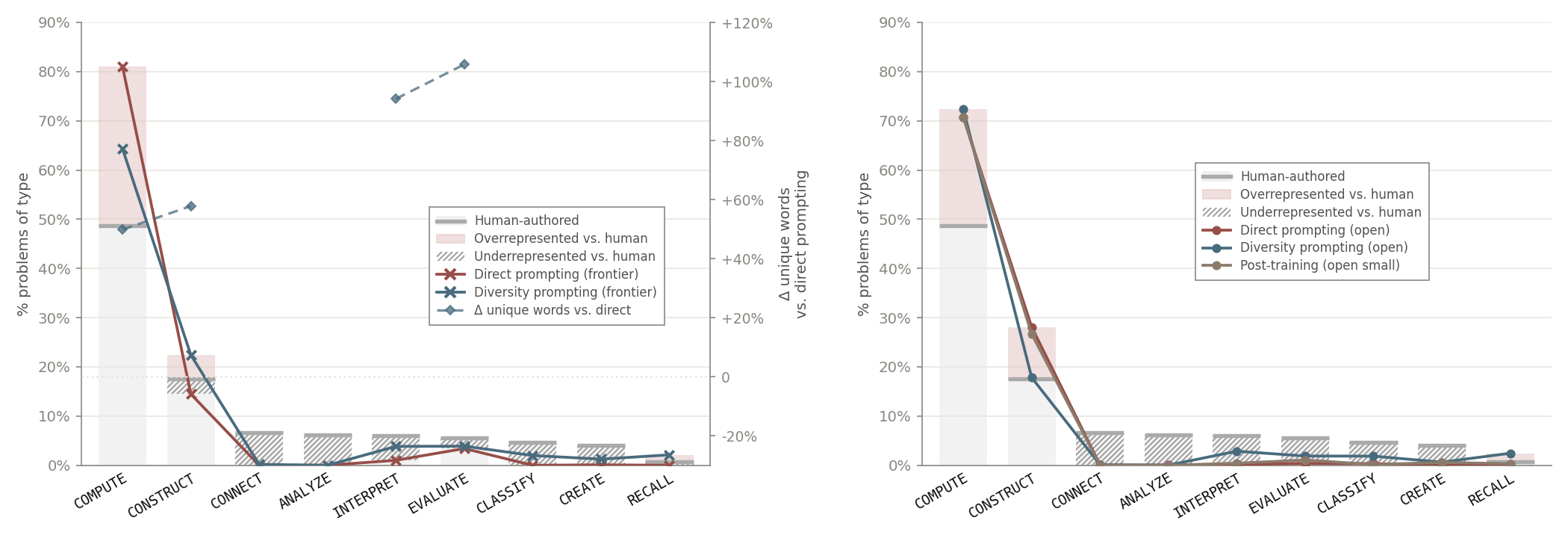}
    \caption{\textbf{Distribution of math task types.} Human-authored and model-generated problems under direct prompting, diversity prompting, and post-training. Left: frontier models; right: open-weight models. Dashed lines show the percentage change in unique words relative to direct prompting (right axis).}
    \label{fig:mathdomain}
    \vspace{-0.3cm}
\end{figure*}

In math problems (Figure~\ref{fig:mathdomain}), \textit{compute} tasks account for approximately 49\% of human-authored problems but 81\% of problems generated through direct prompting. Together, \textit{compute} and \textit{construct} account for approximately 96\% of model-generated problems. Models rarely generate problems requiring students to analyze worked reasoning, connect representations, or create mathematical problems and models. \textit{Analyze} and \textit{connect} each account for less than 1\% of generated problems.

\subsection{Diversity interventions yield limited and superficial gains.}
\label{section:2}
\paragraph{Diversity prompting.}
For frontier models (GPT-5.2, Gemini-3.5-Flash), diversity prompting reduces average Self-BLEU by 33.6\% and embedding similarity by 28.0\%. For open-weight models, prompting has no significant effect on Self-BLEU and reduces embedding similarity by 11.7\%. Despite these gains, frontier model outputs remain 66.2\% more homogeneous than human-authored content on Self-BLEU and 27.3\% on embedding similarity.

\begin{figure*}[h]
    \centering
    \includegraphics[width=\textwidth]{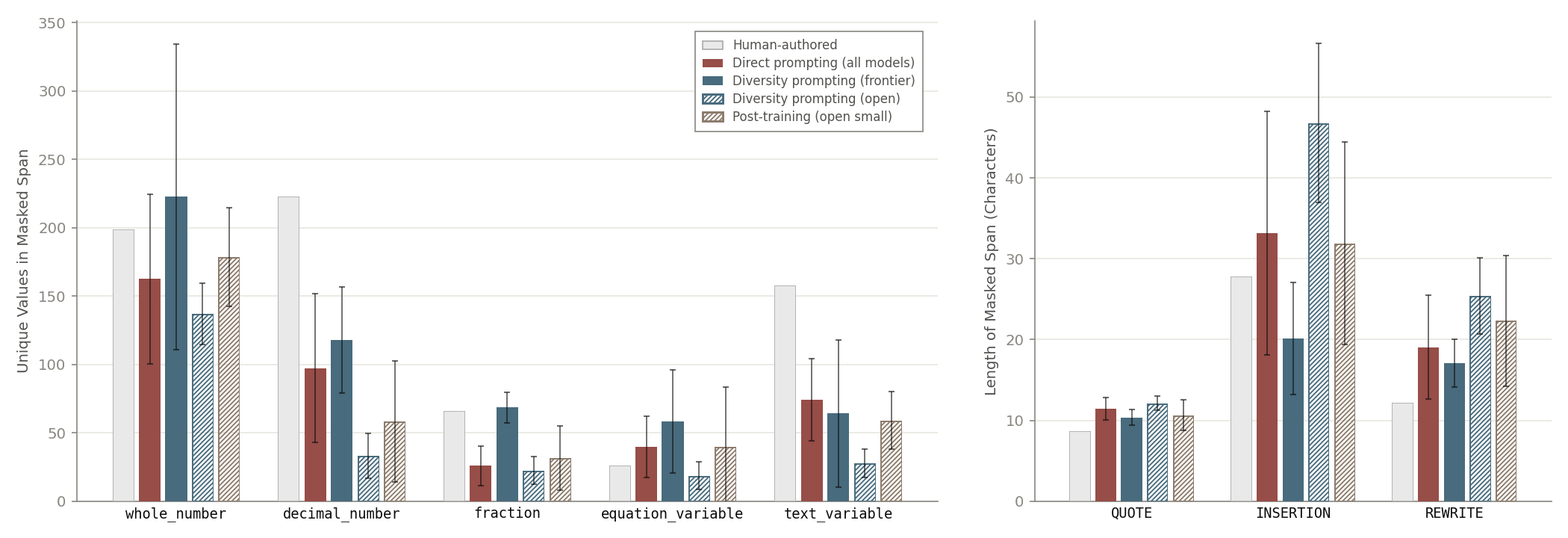}
    \caption{\textbf{Diversity of masked text in math items and essay feedback.} Left: number of unique values for each masked variable type in math problems. Right: average length of masked text spans in essay feedback. Error bars show 95\% confidence intervals across models.}
    \label{fig:masking}
    \vspace{-0.3cm}
\end{figure*}

Some of these gains come from superficial variation. In math, masking removes 25–51\% of the Self-BLEU improvement for frontier models. As shown in Figure~\ref{fig:masking}, diversity prompting increases the variety of numerical values, while human-authored problems use a wider range of contexts and objects. In feedback, models already generate longer suggested insertions and rewrites than human authors. Diversity prompting pushes open-weight models to rely even more on ready-made text, with insertions 65\% longer than human-authored ones and rewrites twice as long.

Compared with direct prompting, diversity prompting increases vocabulary without substantially changing the distribution of instructional forms. In math (Figure~\ref{fig:mathdomain}), \textit{compute} problems decline from 81\% to 64\% but remain overrepresented relative to human-authored problems (49\%), while the underrepresentation of \textit{analyze}, \textit{connect}, and \textit{create} tasks is virtually unchanged. Meanwhile, vocabulary grows by 91\% for \textit{interpret} and 74\% for \textit{evaluate}, although these types account for less than 5\% of generated problems. In feedback (Figure~\ref{fig:feedbackdomain}), \textit{operate} comments and \textit{commands} decrease by 9\% and 8\%, respectively, but both remain overrepresented. Vocabulary increases across nearly every type, including 8\% for \textit{interpret} and 7\% for \textit{open questions}, although both continue to account for less than 5\% of comments.

\paragraph{Post-training.}

SFT reduces average Self-BLEU by 12.6\% and embedding similarity by 20.6\%. DPO reduces Self-BLEU by 10.9\% and embedding similarity by 16.0\%. For smaller open-weight models, post-training improves these conventional diversity metrics more than diversity prompting. However, relative to human-authored content, SFT outputs remain 78.3\% more homogeneous on Self-BLEU and 30.0\% on embedding similarity, and DPO outputs remain 81.9\% and 37.6\% more homogeneous.

Post-training gains are inconsistent across tasks and often come at the expense of acceptability (Appendix Table~\ref{tab:validity}). In math, only 76\% of problems generated by Llama-3.1-8B-Instruct after DPO are solvable and 82\% understandable. For Qwen-2.5-7B-Instruct, DPO offers little improvement over direct prompting. In feedback, DPO training loss collapses almost immediately as stylistic differences between human and model feedback make the preference signal easy to exploit. SFT reduces acceptability, with only 75\% of comments for both models meeting criteria for accuracy or relevance.

Among acceptable outputs, post-training does little to change the distribution of instructional forms. In math (Figure~\ref{fig:mathdomain}), \textit{compute} and \textit{construct} still account for 96\% of post-training problems, leaving just 4\% for the other seven task types. In feedback (Figure~\ref{fig:feedbackdomain}, while DPO largely preserves the distribution observed under direct prompting, SFT learns to overproduce praise and questions: 41.5\% of SFT comments contain questions, compared with 27\% in the human training data. Yet SFT relies on a smaller vocabulary and repeats three-word phrases 3–4 times as often as human comments. It falls back on stock expressions such as ``Why is this relevant?'' that can pass acceptability checks while still being unhelpful or inappropriate in context (Appendix~\ref{appendix:posttrain}).

\section{Discussion}
Despite advances in methods for diversifying LLM outputs~\citep{zhang2025verbalizedsamplingmitigatemode, zhang2026cultivatingpluralismalgorithmicmonoculture}, model-generated educational content remains substantially more homogeneous than human-authored content, both in language and in the instructional practices represented. Reliance on this content restricts the kinds of thinking and disciplinary work students are invited to do. Across both tasks, models amplify routines already common in human-authored materials, crowding out more ambitious forms of instruction. In writing, models favor correcting phrasing or adding more content over asking students to connect ideas, interpret their significance, or evaluate their work~\citep{bereiter1987psychology}. In math, models overwhelmingly produce computation tasks rather than ones that ask students to interpret representations, evaluate claims, or analyze mathematical reasoning~\citep{swan2005improving}. 

As teachers increasingly rely on LLM-generated materials, the instructional repertoire represented by these systems becomes part of how teaching is practiced, understood, and valued. When model-generated materials become widespread, the most easily automated routines become more visible and recognizable as teaching, while teachers have fewer opportunities to exercise and develop the judgment required for more ambitious instruction~\citep{apple1990you, biesta2015education}. As that expertise erodes, the profession itself shrinks, increasingly defined by the work that technology can reproduce.

Developing models that can represent the breadth of human expertise requires investing in teachers' disciplinary knowledge and professional judgment~\citep{darling2006constructing}. Yet even expert teachers rely on routine practices when they are appropriate, and training on their demonstrations encourages models to reproduce patterns of expression rather than the judgment behind instructional decisions. Future work should investigate how models can learn this adaptive expertise~\citep{lin2018toward}, enabling pedagogical diversity through flexible judgment rather than optimizing output distributions.

\section*{Acknowledgments}
This work was supported by Overdeck Family Foundation and the Stanford Institute for Human-Centered Artificial Intelligence (HAI). We thank Jim Malamut and Kristyn Wilson for their expertise in developing the domain measures and codebooks. We are grateful to all the teachers whose collaboration and expertise have supported this work over the years.

\bibliography{colm2026_conference}
\bibliographystyle{colm2026_conference}

\clearpage
\appendix

\section{Source Data}
\label{appendix:source_data}

Tables~\ref{tab:context_examples} and \ref{tab:feedback_context_examples} show examples of the source context provided to models during generation. For math items, this context consisted of the Illustrative Mathematics unit description summarizing the target learning objectives. For feedback, it consisted of the writing assignment and the full student essay.

\begin{table*}[t]
\centering
\small
\setlength{\tabcolsep}{6pt}
\renewcommand{\arraystretch}{1.3}
\begin{tabular}{@{}p{0.19\textwidth}p{0.79\textwidth}@{}}
\toprule
\textbf{Unit ID} & \textbf{Unit Description} \\
\midrule
\textbf{HS\_2\_6} \newline (High School Geometry Unit 6) &
This unit brings together students' experience from previous years with their new understanding from this course for an in-depth study of coordinate geometry. Students encounter a new coordinate transformation notation such as $(x, y) \rightarrow (x+3, y+1)$, which connects transformations to functions. Then they use transformations and the Pythagorean Theorem to build equations of circles, parabolas, parallel lines, and perpendicular lines from definitions. Students apply these ideas to proofs, such as classifying quadrilaterals. Finally, students use weighted averages to partition segments, scale figures, and locate the intersection point of the medians of a triangle. \\
\midrule
\textbf{MS\_1\_4} \newline (Grade 6 Unit 4) &
In this unit, students examine how the relative sizes of numerator and denominator affect the size of their quotient when numerator or denominator (or both) is a fraction. They acquire the understanding that dividing by $a/b$ has the same outcome as multiplying by $b$, then by $1/a$. They compute quotients of fractions. They solve problems involving lengths and areas of figures with fractional side lengths and extend the formula for the volume of a right rectangular prism to prisms with fractional edge lengths and use it to solve problems. They use tape diagrams, equations, and expressions to represent situations involving partitive or quotitive interpretations of division with fractions. Given a multiplication or division equation or expression with fractions, they describe a situation that it could represent. They use tape diagrams and equations in reasoning about situations that involve multiplication and division of fractions. \\
\midrule
\textbf{MS\_2\_5} \newline (Grade 7 Unit 5) &
In this unit, students interpret signed numbers in contexts (e.g., temperature, elevation, deposit and withdrawal, position, direction, speed and velocity, percent change) together with their sums, differences, products, and quotients. (``Signed numbers'' include all rational numbers, written as decimals or in the form $a/b$.) Students use tables and number line diagrams to represent sums and differences of signed numbers or changes in quantities represented by signed numbers such as temperature or elevation, becoming more fluent in writing different numerical addition and subtraction equations that express the same relationship. They compute sums and differences of signed numbers. They plot points in the plane with signed number coordinates, representing and interpreting sums and differences of coordinates. They view situations in which objects are traveling at constant speed (familiar from previous units) as proportional relationships. For these situations, students use multiplication equations to represent changes in position on number line diagrams or distance traveled, and interpret positive and negative velocities in context. They become more fluent in writing different numerical multiplication and division equations for the same relationship. Students extend their use of the ``next to'' notation (which they used in expressions such as $5x$ and $6(3+2)$ in grade~6) to include negative numbers and products of numbers, e.g., writing $-5x$ and $(-5)(-10)$ rather than $(-5) \cdot (x)$ and $(-5) \cdot (-10)$. They extend their use of the fraction bar to include variables as well as numbers, writing $-8.5 / x$ as well as $\frac{-8.5}{x}$. \\
\bottomrule
\end{tabular}
\caption{Example unit descriptions provided to models as generation context.}
\label{tab:context_examples}
\end{table*}
\begin{table*}[t]
\centering
\small
\setlength{\tabcolsep}{6pt}
\renewcommand{\arraystretch}{1.3}
\begin{tabular}{@{}p{0.32\textwidth}p{0.64\textwidth}@{}}
\toprule
\textbf{Assignment Prompt} & \textbf{Student Essay} \\
\midrule
Read ``Narciso Rodriguez'' by Narciso Rodriguez. Describe the mood created by the author in the memoir. Support your answer with relevant and specific information from the memoir. &
The mood in the memoir that the author creates is a very warm and relaxed sort of mood. The way the author tells you about his parents; the selflessness and the way they put their kids first before themselves really just makes you think about your own parents and how good they are to you and what they would do for you. The way the author talks about his home makes you think of all the colors in the house. I think that the house would be a golden red color yellow, white, brown and orange on the insides. It reminds me of the way they would have cooked. They would have cooked with colorful peppers, onions, tortillas and oils. It makes you think about what colors you have in your home and what you would cook with. The music sounds wonderful. Even though they don't demonstrate what it actually sounds like, you can easily imagine it. I imagine it laying in the background while all the adults are cooking and the kids are playing in colorful rooms with the golden sunlight shining in through an open window. I love the mood that this memoir puts me in. \\
\midrule
Your principal is reconsidering the school's cell phone policy, choosing between two options. Policy~1: allow phones at school for use during lunch and free times, turned off during class. Policy~2: do not allow phones at school at all. Write a letter to your principal convincing her which policy is better. Support your position with specific reasons. &
Dear Principal, I have heard about the new cell phone policies that you have been thinking of. I feel that policy number 2 is the better one of the 2 choices. hear are a couple of the reasons why I personally think this. Students enjoy using their phone every day. I feel that sometimes the can use it a little to much during school. I think that letting students use their phones during lunch and free time is an excellent idea. Also students wouldn't be interrupting class every day with their cell phones. This would be a lot better than just getting rid of cell phones completely. As principal you might want to consider possibly figuring out a schedule were students have free time during a couple of their classes. Doing this would not only help the new policy, but keep it going throughout the school year. students would appreciate it if during a boring class they got a 5-15 minute break to relax and use their phones. But, there would be some rules behind the free time. You would have to think of them so that students wouldn't be doing stuff that is against school rules. So principal please consider the second policy for cell phones at our school. students do want to use their cell phones and not keep them at home. I hope you liked the reasons why please se \\
\bottomrule
\end{tabular}
\caption{Example prompt-essay pairs provided to models as generation context.}
\label{tab:feedback_context_examples}
\end{table*}

\section{Masking}
\label{appendix:masking}
To estimate how much of the observed variation is driven by surface-level values, we created masked versions of the math problems and feedback comments. We prompted GPT-5-nano to apply the masks using the prompts shown in Figures~\ref{fig:mask-math-prompt} and \ref{fig:mask-feedback-prompt}. To programmatically narrow the cases requiring human review, we used \texttt{difflib.SequenceMatcher} to align each original text with its masked version and recover the original value replaced by each tag.
\begin{figure}[t]
\centering
\begin{lstlisting}[style=prompt]
You will be shown a math problem. 

Create a maskedProblem by replacing:
- Whole numbers with [whole_number]
- Decimals with [decimal_number]
- Fractions with [fraction]
- Variables in equations, such as x or y, with [equation_variable]
- Names and labels with [text_variable]

Preserve all other wording and structure. Return only the masked problem.

## Example 1
problem: Jada has a sleeping bag that is rated for 30 degrees F.
Write an inequality that represents the outdoor temperature at which
Jada will be able to stay warm.
maskedProblem: [text_variable] has a sleeping bag that is rated for
[whole_number] degrees F. Write an inequality that represents the
outdoor temperature at which [text_variable] will be able to stay warm.

## Example 2
problem: Point $A$ is located at $(-3, 2)$ and point $B$ is
located at $(7, 17)$ on a coordinate plane. Find the coordinates of
point $P$ that partitions the directed line segment $AB$ in a $3:2$
ratio.
maskedProblem: Point [text_variable] is located at ([whole_number],
[whole_number]) and point [text_variable] is located at
([whole_number], [whole_number]) on a coordinate plane. Find the
coordinates of [text_variable] that partitions the directed line
segment between [text_variable] and [text_variable] in a
[whole_number]:[whole_number] ratio.

## Example 3
problem: Write a number puzzle that results in the equation
3x - 5 = 16.
maskedProblem: Write a number puzzle that results in the equation
[whole_number][equation_variable] - [whole_number] = [whole_number].
\end{lstlisting}
\caption{System prompt for masking math problems.}
\label{fig:mask-math-prompt}
\end{figure}

\begin{figure}[t]
\centering
\begin{lstlisting}[style=prompt]
You will be shown teacher feedback on a student essay.
The input is formatted with:
- EXCERPT: a window of essay text with brackets marking the highlighted portion
- COMMENT: the teacher feedback

Create a maskedComment by replacing:
- wording quoted or near-quoted from the EXCERPT with <QUOTE>
- replacement wording for something in the EXCERPT with <REWRITE>
- new wording the teacher proposes adding to the essay with <INSERTION>

Use the EXCERPT to distinguish quotes, rewrites, and insertions. Mask the smallest meaningful span and preserve all other wording and structure. Remove any quotation marks surrounding the masked span. Return only the maskedComment.

## Example 1
EXCERPT: Driverless cars could help people.
COMMENT: Clarify this idea by saying something like, "One good thing
about driverless cars is that they could help elderly people."
maskedComment: Clarify this idea by saying something like, <INSERTION>.

## Example 2
EXCERPT: Their are more disadvantages because the driver fins more
problems then expected.
COMMENT: Correct the spelling of 'their' to 'there', 'fins' to
'finds', and 'then' to 'than'.
maskedComment: Correct the spelling of <QUOTE> to <REWRITE>, <QUOTE>
to <REWRITE>, and <QUOTE> to <REWRITE>.

## Example 3
EXCERPT: The mood creates unity and a feeling of togetherness.
COMMENT: How does 'unity' differ from or align with your initial
description of the mood as 'togetherness'?
maskedComment: How does <QUOTE> differ from or align with your initial
description of the mood as <QUOTE>?

## Example 4
EXCERPT: Over all I see benfit in allowing driverless cars.
COMMENT: Use 'Overall, I see benefits' instead of 'Over all I see
benfit' - make sure to proofread for typos and grammatical errors.
maskedComment: Use <REWRITE> instead of <QUOTE> - make sure to
proofread for typos and grammatical errors.

## Example 5
EXCERPT: Long-distance driving can have physical effects.
COMMENT: Consider using more formal or technical language to describe
the physical effects of long-distance driving, such as 'fatigue' or
'muscle strain'.
maskedComment: Consider using more formal or technical language to
describe the physical effects of long-distance driving, such as
<INSERTION> or <INSERTION>.
\end{lstlisting}
\caption{System prompt for masking teacher feedback comments.}
\label{fig:mask-feedback-prompt}
\end{figure}

\paragraph{Math items.}
We manually audited cases in which the extraction procedure either failed to recover a value for a tag or recovered a value that did not match the expected tag type, such as a nondigit value for a whole-number tag. A common source of extraction failures was adjacent tags, as in “line AB” when A and B had previously been introduced as separate points. Of the 1,668 cases flagged for this issue, 1,200 were resolved programmatically using stricter regular expressions for the character types associated with each tag. Of the remaining 468 cases, we manually corrected three masking errors. Extraction also failed when model outputs included process text or formatting labels before the masked problem, such as “Wait, I need to ensure correct formatting” or “\#\# Anonymized text:”. We audited 749 of such cases and corrected 69 corrupted outputs.

\paragraph{Essay feedback.}
For feedback comments, we manually audited 56 cases in which extraction failed and 99 cases in which masking replaced the entire comment. We also flagged 888 cases with empty extracted values and 956 cases with unmatched quotation marks. All 888 empty values and 25 of the unmatched quotation cases reflected extraction errors, which we corrected programmatically. For the remaining 168,441 comments, we ranked outputs by character overlap between the original and masked versions and manually reviewed the 1,011 comments with less than 30\% overlap. Across 3,010 audited comments, we found and manually corrected two masking errors.

\section{Acceptability}
\label{appendix:acceptability}
To establish a baseline level of quality, we assessed the acceptability of each output before conducting the diversity analyses. We used a panel of three LLM judges, GPT-5-nano, Claude-Haiku-4.5, and Gemini-2.5-Flash-Lite, using the prompts shown in Figures~\ref{fig:math-prompt} and \ref{fig:feedback-prompt}. For math problems, judges rated understandable language, solvability, and relevance to the unit. For feedback comments, they rated understandable language, accuracy, and helpfulness. We report acceptability ratings in Table~\ref{tab:validity}. We pooled the three criteria within each judgment, treating an output as invalid if the judge assigned a 0 to any criterion.

\begin{table*}[t]
\centering
\small
\setlength{\tabcolsep}{4pt}
\begin{tabular}{@{}l|ccc|ccc@{}}
& \multicolumn{3}{c|}{\bf Math Items} & \multicolumn{3}{c}{\bf Essay Feedback} \\
\textbf{Generator} & \textbf{Underst.} & \textbf{Solvable} & \textbf{Unit Rel.} & \textbf{Underst.} & \textbf{Accurate} & \textbf{Context Rel.} \\
\midrule
\textbf{Human} & \textbf{99.1} & \textbf{98.3} & \cellcolor{darkred!25}\textbf{85.4} & \cellcolor{darkred!10}\textbf{91.3} & \cellcolor{darkred!17}\textbf{88.6} & \cellcolor{darkred!18}\textbf{88.2} \\
\midrule
\textbf{GPT-5.2} & \textbf{99.9} & \textbf{99.9} & \textbf{99.9} & \textbf{100.0} & \textbf{100.0} & \textbf{100.0} \\
\quad + Verbalized sampling & 100.0 & 99.9 & 99.6 & 100.0 & 100.0 & 100.0 \\
\quad + Negative sampling & 99.9 & 99.9 & 99.2 & 100.0 & 100.0 & 100.0 \\
\textbf{Gemini-3.5-Flash} & \textbf{100.0} & \textbf{100.0} & \textbf{99.9} & \textbf{100.0} & \textbf{100.0} & \textbf{100.0} \\
\quad + Verbalized sampling & 99.8 & 99.7 & 99.3 & 100.0 & 100.0 & 100.0 \\
\quad + Negative sampling & 99.8 & 99.6 & 98.9 & 100.0 & 100.0 & 100.0 \\
\textbf{Llama-3.3-70B} & \textbf{98.6} & \textbf{97.9} & \textbf{98.3} & \textbf{99.5} & \textbf{95.3} & \textbf{95.5} \\
\quad + Verbalized sampling & 98.0 & 96.1 & \cellcolor{darkred!3}94.0 & 97.2 & \cellcolor{darkred!18}88.2 & \cellcolor{darkred!18}88.3 \\
\quad + Negative sampling & 98.7 & 97.4 & \cellcolor{darkred!3}94.0 & 97.0 & \cellcolor{darkred!17}88.4 & \cellcolor{darkred!17}88.5 \\
\textbf{Qwen-3.6-27B} & \textbf{99.3} & \textbf{98.2} & \textbf{99.3} & \textbf{98.3} & \cellcolor{darkred!18}\textbf{88.3} & \cellcolor{darkred!15}\textbf{89.2} \\
\quad + Verbalized sampling & 99.4 & 98.1 & 97.5 & \cellcolor{darkred!3}94.0 & \cellcolor{darkred!52}75.2 & \cellcolor{darkred!49}76.4 \\
\quad + Negative sampling & 99.3 & 98.8 & 98.3 & 96.4 & \cellcolor{darkred!42}78.9 & \cellcolor{darkred!39}80.2 \\
\textbf{Llama-3.1-8B-Instruct} & \textbf{96.2} & \cellcolor{darkred!6}\textbf{92.6} & \cellcolor{darkred!9}\textbf{91.6} & \textbf{98.9} & \cellcolor{darkred!17}\textbf{88.4} & \cellcolor{darkred!14}\textbf{89.7} \\
\quad + Verbalized sampling & 98.6 & 95.4 & \cellcolor{darkred!12}90.6 & 96.1 & \cellcolor{darkred!26}85.2 & \cellcolor{darkred!24}86.0 \\
\quad + Negative sampling & \cellcolor{darkred!2}94.1 & \cellcolor{darkred!9}91.5 & \cellcolor{darkred!11}90.8 & 95.8 & \cellcolor{darkred!28}84.4 & \cellcolor{darkred!25}85.3 \\
\quad + SFT & 98.1 & \cellcolor{darkred!1}94.8 & \cellcolor{darkred!11}90.9 & \cellcolor{darkred!4}93.6 & \cellcolor{darkred!52}75.0 & \cellcolor{darkred!52}75.3 \\
\quad + DPO & \cellcolor{darkred!35}81.8 & \cellcolor{darkred!50}76.1 & \cellcolor{darkred!34}81.9 & 98.0 & \cellcolor{darkred!35}81.8 & \cellcolor{darkred!29}83.8 \\
\textbf{Qwen-2.5-7B} & \textbf{99.2} & \textbf{98.1} & \textbf{98.0} & \textbf{97.9} & \cellcolor{darkred!33}\textbf{82.3} & \cellcolor{darkred!29}\textbf{83.8} \\
\quad + Verbalized sampling & 99.3 & 98.2 & \cellcolor{darkred!2}94.4 & 97.3 & \cellcolor{darkred!39}80.2 & \cellcolor{darkred!34}82.1 \\
\quad + Negative sampling & 97.7 & 96.7 & \cellcolor{darkred!4}93.4 & 97.7 & \cellcolor{darkred!35}81.6 & \cellcolor{darkred!32}82.9 \\
\quad + SFT & 99.0 & 98.3 & 97.4 & \cellcolor{darkred!7}92.2 & \cellcolor{darkred!52}75.1 & \cellcolor{darkred!55}74.0 \\
\quad + DPO & 98.0 & 95.8 & 95.9 & 97.5 & \cellcolor{darkred!36}81.1 & \cellcolor{darkred!33}82.4 \\
\bottomrule
\end{tabular}
\caption{\textbf{Validity of generated items across math problems and essay feedback.} Percentage of generated items satisfying each quality criterion. Higher values indicate greater validity; cell shading marks values below 95\%, with darker shading denoting lower validity.}
\label{tab:validity}
\end{table*}

\begin{figure}[t]
\centering
\begin{lstlisting}[style=prompt]
You will be shown a math problem.
The input is formatted with:
* GRADE_LEVEL: the intended grade level
* LESSON_NARRATIVE: the lesson goals and content
* MATH_PROBLEM: the problem students would see

Evaluate whether the MATH_PROBLEM valid. Score it on these 3 binary criteria:

1. understandable_language
* 1 = the problem is clear, readable, and complete enough to understand
* 0 = the problem is broken, confusing, malformed, or too incomplete to
      interpret

2. solvable
* 1 = the problem has at least one valid answer as written
* 0 = it is missing necessary information, contradictory, mathematically
      invalid, or too ambiguous to solve
Solve the problem before scoring it. Open-ended problems are solvable if
there are acceptable answers.

3. unit_relevant
* 1 = the mathematical work required fits the GRADE_LEVEL and
      LESSON_NARRATIVE
* 0 = it assesses unrelated content or is inappropriate for the intended
      grade level
Use both the problem and its solution to judge relevance.

Return ONLY valid JSON:
{"understandable_language": 0, "solvable": 0, "unit_relevant": 0}
\end{lstlisting}
\caption{System prompt for the math-problem validity judges.}
\label{fig:math-prompt}
\end{figure}

\begin{figure}[t]
\centering
\begin{lstlisting}[style=prompt]
You will be shown feedback on a student essay.
The input is formatted with:
- ASSIGNMENT: the writing task or prompt the student was responding to
- FULL_ESSAY: the student's full essay
- EXCERPT: a window of essay text with brackets marking the highlighted
  portion
- COMMENT: the feedback comment

Evaluate whether the COMMENT is valid feedback on the student's writing. 
Score it on these 3 binary criteria:

1. understandable_language
* 1 = the comment is understandable natural language and functions as
      interpretable feedback in relation to the excerpt
* 0 = the comment is gibberish, broken, truncated, malformed, cannot be
      understood as feedback on the excerpt, is unrelated to the excerpt,
      or consists of unrelated fragments

2. accurate
* 1 = the comment correctly assesses or responds to     
    something in the writing and any proposed revision is valid
* 0 = the comment misreads the essay, invents or 
    misdiagnoses an issue, or proposes a false, identical, contradictory, or inappropriate correction

3. context_relevant
* 1 = the comment makes a meaningful suggestion or 
    observation that improves the student's writing or supports the student's development as a writer.
* 0 = the comment makes no meaningful request or 
    observation, is off-topic or directs the student toward unproductive revisions

Return ONLY valid JSON:
{"understandable_language": 0, "accurate": 0, "context_relevant": 0}
\end{lstlisting}
\caption{System prompt for the essay-feedback validity judges.}
\label{fig:feedback-prompt}
\end{figure}

\paragraph{Math items.} We manually reviewed the 1,616 outputs for which the three judges did not unanimously agree. Human review agreed with the majority judgment in 73.8\% of these cases and with the dissenting judgment in 26.2\%. 

Through the audit, we discovered that judges accepted invalid problems with contradictory language, multiple-choice questions without the correct answer among the options, problems with references to missing diagrams, tables, or previous problems, and other malformed generations that repeated sub-questions or ended abruptly mid-sentence. We also found that judges rejected some valid problems, including simple interpretation questions, such as identifying a unit rate or writing an equation, answerable questions containing "distractor" information, and questions with valid edge-case answers, such as an undefined slope. Notably, only 85\% of the human-authored math problems were judged as unit-relevant. We manually reviewed all 79 human-authored problems flagged as invalid for this reason and found that they were indeed review problems included in the source curriculum.

\paragraph{Essay feedback.} Because judgments about accuracy and helpfulness comprised 95\% of disagreements, we treated comments flagged by any judge for understandable language as invalid. We retained unanimous judgments and marked the remaining 37,056 comments for further review. Reviewing all comments by hand was not feasible, so we randomly sampled 200 disagreements to identify common sources of ambiguity and incorporated these patterns into an adjudication prompt for Claude-Sonnet-5. The adjudicator agreed with the minority judgment in 36.4\% of cases to overturn a majority valid judgment and 28.5\% of cases to overturn a majority invalid judgment. We manually reviewed a sample of 5,000 cases in which the adjudicator overturned the majority judgment and agreed with the adjudicator in 96\% of cases. 

Through the audit, we discovered that judges accepted feedback that called something missing or incorrect when it was present and correct elsewhere in the essay, feedback that misdiagnosed an issue, feedback that suggested content that contradicted the student’s ideas. We also found that judges rejected valid comments that were vague but still grounded in the student's writing, as well as feedback referencing external facts or sources that could not be verified from the essay but were not demonstrably false. Notably, only 88\% of human-authored feedback comments were judged as accurate and helpful. We manually reviewed all 448 human-authored comments flagged as invalid for this reason. Human comments were often penalized for forms of feedback that were less common in model outputs. These included praise that served a relational or personal function, such as “I love that you stay in contact with the library. This is so cute,” and short but legitimate corrections, such as “comma after to". Judges also rejected comments that correctly identified contradictions or problems visible only from the full essay rather than the selected excerpt. These included comments that redirected the student to the assignment or pointed out that the student’s current claim conflicted with an earlier claim. At the same time, the human data included genuinely invalid comments, such as “same comment as earlier,” and corrections too abbreviated to interpret without additional context.

\section{Post-training}
\label{appendix:posttrain}

We implemented SFT and DPO using OpenRLHF~\citep{hu2024openrlhf}. Models were loaded in \texttt{bfloat16}. For SFT, human-authored outputs served as training targets. For DPO, human-authored outputs served as chosen responses and were paired with model-generated rejected responses.

For math items, we fine-tune using SFT for 2 epochs with a learning rate of ($5\times10^{-7}$) and a batch size of 32. For DPO, we train for 1 epoch with a learning rate of ($5\times10^{-7}$), batch size of 32, and $\beta$ coefficient 0.1. We use bfloat 16 precision for both and train until convergence.

For essay feedback, we used a learning rate of \(1 \times 10^{-5}\) and trained for two epochs for SFT, as the smaller dataset benefited from additional training. For DPO, we initially tested a learning rate of \(1 \times 10^{-6}\), \(\beta = 0.2\), and two epochs, but these settings produced unstable training and malformed outputs. We ultimately used a learning rate of \(5 \times 10^{-7}\) and trained for one epoch.

\begin{figure*}[h]
    \centering
    \includegraphics[width=\textwidth]{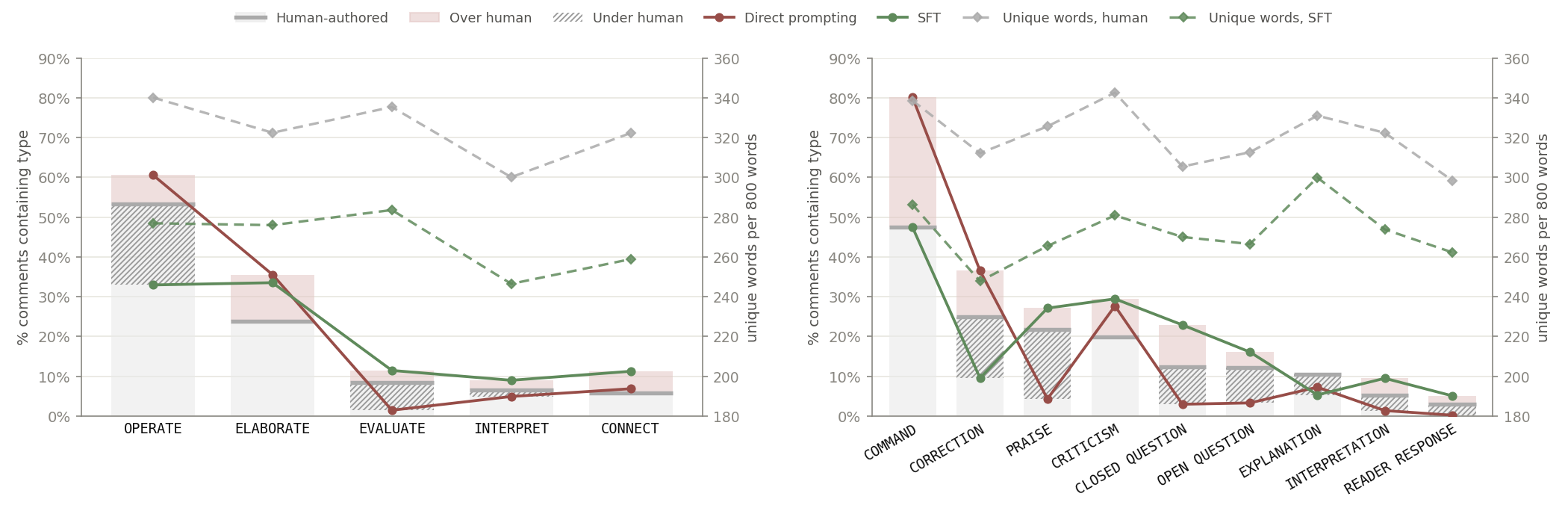}
    \caption{\textbf{Effects of SFT on essay feedback.} Distribution of feedback revision types (left) and rhetorical strategies (right) for human-authored feedback, direct prompting, and SFT. Dashed lines show lexical diversity (unique words in equal-length samples, right axis).}
    \label{fig:feedbackdomainsft}
    \vspace{-0.3cm}
\end{figure*}

Though post-training generally yielded limited improvements in pedagogical diversity, SFT on essay feedback was an exception. As shown in Figure~\ref{fig:feedbackdomainsft}, SFT copies human-authored expressions and overshoots prevalence for several strategies, including closed questions (23\% vs. 12.5\%), criticism (29\% vs. 19\%), and praise (27\% vs. 22\%). In fact, 41.5\% of SFT comments contain questions, compared with 27\% of human comments in the training data.

Yet this comes at the expense of lexical diversity. Within each type, SFT draws on a smaller, more repetitive vocabulary. It produces significantly fewer unique words than human feedback across every type, controlling for text length, and repeats three-word phrases 3–4 times as often. SFT appears to achieve its broader rhetorical coverage through recurring expressions such as ``Why is this relevant?'', ``This looks like your claim,'' ``Great start!,'' and ``Is this necessary?,'' which can pass baseline acceptability checks while remaining unhelpful or inappropriate in context. SFT reproduces surface patterns of expression without the pedagogical judgment that guides their use.

\section{Domain Measures}
\label{appendix:domainmodels}
We developed domain-grounded codebooks for Math Task Type, Feedback Revision Type, and Feedback Rhetorical Strategy, presented in Tables~\ref{tab:math_task_codebook}, \ref{tab:feedback_codebook}, and \ref{tab:feedback_move_codebook}. We manually examined 50 randomly sampled examples from each task and found that most math problems could be assigned a single task type, whereas feedback comments frequently combined multiple revision requests or rhetorical strategies. We therefore use single-label classification for math items and multi-label classification for both feedback measures.

We independently classified outputs using Claude-Sonnet-5 and Gemini-3.1-Flash-Lite, with the prompts shown in Figures~\ref{fig:math-llm-judge-prompt}, \ref{fig:feedback1-llm-judge-prompt} and \ref{fig:feedback2-llm-judge-prompt}. The judges agreed on 90\% of math task labels, 76\% of feedback revision labels, and 68\% of feedback rhetorical strategy labels (Cohen's $\kappa$ values of 0.78, 0.72 and 0.70, respectively, indicating substantial agreement). 

Reviewing judge disagreements, we found that Gemini made more classification errors than Claude. Gemini's errors included clear misapplications of the codebooks; for example, labeling Construct for a word problem requiring computation to get an answer or Open-Question for a general command to explain the significance of a quote. Other disagreements reflected genuine ambiguities, with both models relying on wording cues to resolve them. For example, they used Interpret for requests to ``explain why'' even when students were simply asked to elaborate, or Evaluate for judgments that writing could be ``stronger'' or ``clearer'' that did not invite students to assess their own work. In math, models overused Recall for questions asking students to interpret information provided in the problem, such as identifying the domain and range of a function. They also labeled Analyze for questions with a lengthy mathematical setup, even though the task was to ``select all true statements''.

We therefore iteratively reviewed 200 examples from each task, sampled across label categories, and corrected Claude's classification errors by adding instructions and examples until it correctly labeled all 200 examples for each task. Finally, we manually labeled a new stratified random sample of 100 outputs per task, achieving 88\% agreement with Claude for Math Task Type, 96\% for Feedback Revision Type, and 92\% for Rhetorical Strategy.

\begin{table*}[t]
\centering
\small
\setlength{\tabcolsep}{6pt}
\renewcommand{\arraystretch}{1.3}
\begin{tabular}{@{}p{0.11\textwidth}p{0.46\textwidth}p{0.37\textwidth}@{}}
\toprule
\textbf{Code} & \textbf{Definition} & \textbf{Example} \\
\midrule
\textbf{Recall} & Retrieving a fact, definition, formula, or convention. Nothing is produced nor examined. & What is the appropriate unit for volume: cm, cm$^2$, or cm$^3$? \\
\midrule
\textbf{Interpret} & Surfacing or reasoning about meaning: what a mathematical object, symbol, or quantity refers to or implies about a situation; showing conceptual understanding, not carrying out a procedure or producing a representation. & If the 5 represents the number of people, what does the 4 represent? \\
\midrule
\textbf{Compute} & Finding a numerical, symbolic, or exact answer: calculating, simplifying, solving, or determining a value. & Solve $3x + 5 = 20$ for $x$. \newline A shirt costs \$40 and is 25\% off. Find the sale price. \\
\midrule
\textbf{Construct} & Producing a mathematical representation, such as a graph, table, diagram, dot plot, number line, expression, equation, model, or visual display. Lower-level than Create, more routine and less inventive. & Make a dot plot for a given list of data values. \newline Write an expression to represent the perimeter of this figure. \\
\midrule
\textbf{Create} & Generating something new and open-ended where many results are valid: inventing an example, data set, problem, representation, or situation that meets given conditions. More inventive and less routine than Construct. & Create a data set with a mean of 10 and a median of 8. \newline Create a problem that can be solved with $3/4 \div 1/2$. \\
\midrule
\textbf{Classify} & Organizing objects by a property: sorting into groups or categories, labeling, or ordering along a dimension. & Group expressions according to whether they are linear, quadratic, or exponential. \newline Order from smallest to largest: number of pennies in a stack 1\,ft high, number of books in a stack 1\,ft high, \ldots \\
\midrule
\textbf{Connect} & Engaging with how different representations express the same mathematical object, quantity, relationship, or structure; matching, comparing, recognizing similarity. (Use when this is the point of the item, not when contending with multiple representations is a vehicle for solving something else.) & Select all expressions equivalent to $16x + 36$. \newline Would this information be better represented using a table or a double number line? \\
\midrule
\textbf{Evaluate} & Judging whether a mathematical statement, claim, conjecture, or generalization holds: true or false, sometimes/always/never, or whether it fits a situation. & Is the sum of two rational numbers always rational? \newline Are $2b+b$ and $3b$ equivalent for all values of $b$? \\
\midrule
\textbf{Analyze} & Examining mathematical reasoning, a worked solution, a strategy, an argument, a proof, an error, or the results a procedure produced. Includes comparing strategies, critiquing reasoning, debugging mistakes, explaining errors, or deciding whether a solution method is valid. & A student says $4(x+3) = 4x+3$; explain the mistake. \newline Lin measured a diameter vertically (3.5\,cm) and horizontally (3.6\,cm); give possible reasons the measurements differ. \\
\bottomrule
\end{tabular}
\caption{Codebook for classifying the mathematical \texttt{Task Type}}
\label{tab:math_task_codebook}
\end{table*}

\begin{table*}[t]
\centering
\small
\setlength{\tabcolsep}{6pt}
\renewcommand{\arraystretch}{1.3}
\begin{tabular}{@{}p{0.11\textwidth}p{0.46\textwidth}p{0.37\textwidth}@{}}
\toprule
\textbf{Code} & \textbf{Definition} & \textbf{Examples} \\
\midrule
\textbf{Operate} & The feedback asks the student to retrieve, add, correct, or execute something straightforward. The remedy is clear and does not require judgment, reconsideration, or new reasoning. & Fix the spelling of ``necessary.'' \newline Add a citation for this statistic. \newline Start a new paragraph here. \\
\midrule
\textbf{Elaborate} & The feedback asks the student to add more content, detail, explanation, context, or development of a kind they already understand and could produce. The student elaborates without needing to reconsider their claim, belief, purpose, or rhetorical plan. & Can you give an example here? \newline Add something to connect these paragraphs. \newline Your introduction needs more context. \newline Expand on this point. \\
\midrule
\textbf{Connect} & The feedback asks the student to work out relationships between ideas, evidence, claims, structure, counterarguments, or rhetorical choices. The student must construct a connection that is missing, unclear, contradictory, or underdeveloped. & How does this evidence support this claim? \newline You have two contradictory points here. How do they fit together? \\
\midrule
\textbf{Interpret} & The feedback asks the student to develop an idea by explaining its meaning, significance, implication, or application. The student goes beyond adding more content to show what the idea means or why it matters. & What can we infer about her feelings from this quote? \newline Where can we see this same phenomenon in action today? \\
\midrule
\textbf{Evaluate} & The feedback asks the student to reconsider validity, credibility, appropriateness, effectiveness, or fit. The student must assess whether a claim, source, choice, stance, tone, word, or rhetorical move works. & What might a critic say in response to this argument? \newline Is this the right tone for this audience? \newline I'm not sure this is true. \\
\bottomrule
\end{tabular}
\caption{Codebook for classifying the feedback \texttt{Revision Type}.}
\label{tab:feedback_codebook}
\end{table*}

\begin{table*}[t]
\centering
\small
\setlength{\tabcolsep}{6pt}
\renewcommand{\arraystretch}{1.3}
\begin{tabular}{@{}p{0.11\textwidth}p{0.46\textwidth}p{0.37\textwidth}@{}}
\toprule
\textbf{Code} & \textbf{Definition} & \textbf{Examples} \\
\midrule
\textbf{Correction} & The teacher directly fixes something in the text, supplies replacement wording, or gives an explicit edit to the student's text. & Comma splice. \newline Should be ``their'' not ``there.'' \newline Change X to Y. \\
\midrule
\textbf{Command} & The teacher directs the student to take an action, without supplying the exact edit. This includes imperatives and softened directives like ``try,'' ``consider,'' or ``you might.'' & Explain why\ldots \newline Add a transition here. \newline Consider adding more detail. \newline Try to develop this argument further. \\
\midrule
\textbf{Criticism} & The teacher evaluates the writing negatively or identifies a problem. The teacher is judging the writing, even if the criticism is hedged. & This is unclear. \newline These arguments are not convincing. \newline I'm not sure this argument works. \\
\midrule
\textbf{Praise} & The teacher evaluates the writing positively. Includes simple praise and elaborated praise. & Good. \newline Strong argument here. \newline You've defined the issues clearly. \\
\midrule
\textbf{Interpretation} & The teacher neutrally describes, summarizes, or plays back what the student did in the text, without evaluating it or asking for a change. & You're arguing here that legalization would reduce crime. \newline In your conclusion you return to the opening question. \\
\midrule
\textbf{Explanation} & The teacher provides information, reasoning, instruction, a lesson, or a general principle. The comment explains why something matters or teaches the student something. & Transitions help readers follow the logic between your ideas. \newline Direct quotes need quotation marks because readers need to know whose words these are. \\
\midrule
\textbf{Closed Question} & The teacher asks a question that implies an evaluation, problem, or expected answer. The teacher's perspective is implicit in the question. & Have you proved this point by just saying it is so? \newline Can you break up these sentences so I can follow your ideas more easily? \\
\midrule
\textbf{Open Question} & The teacher asks a question that genuinely opens a problem for the student to think through. The student has real latitude in how to respond. & What is it that you want me most to know about your position? \newline How can you explain and support these views? \\
\midrule
\textbf{Reader Response} & The teacher shares their experience as a reader --- what they noticed, felt, understood, wondered, or struggled with --- without directly judging the writing or directing a change. & I found myself confused here. \newline I can tell you feel strongly about this subject. \newline I wonder what you gain by having this paragraph here. \\
\bottomrule
\end{tabular}
\caption{Codebook for classifying the \texttt{Rhetorical Strategy} used to deliver feedback.}
\label{tab:feedback_move_codebook}
\end{table*}

\begin{figure}[t]
\centering
\begin{lstlisting}[style=prompt]
You are an expert math-education researcher coding math problems by the type of cognitive work the problem asks the student to do.

Apply these rules:
Read the problem, decide which single code best captures the primary work the student is asked to do, and call the `assign_code` function with ONLY the code id -- no explanation, no extra text. If a problem seems to span two codes, pick the one that captures its PRIMARY demand -- the thing the student's final answer/response actually consists of.

Each problem gets EXACTLY ONE code from this fixed set:
\{
Label: 
Definition: 
Example: 
\}
\end{lstlisting}
\caption{System prompt for labeling math item task type.}
\label{fig:math-llm-judge-prompt}
\end{figure}

\begin{figure}[t]
\centering
\begin{lstlisting}[style=prompt]
You are an expert education researcher coding essay feedback by the type of revision work the comment asks the student to do. The input is formatted with:
- EXCERPT: a window of essay text with brackets marking the highlighted portion
- COMMENT: the teacher feedback

Select exactly ONE code that best characterizes this comment overall.

## Codebook:
\{
Label: 
Definition: 
Example: 
\}
\end{lstlisting}
\caption{System prompt for labeling essay feedback revision type.}
\label{fig:feedback1-llm-judge-prompt}
\end{figure}

\begin{figure}[t]
\centering
\begin{lstlisting}[style=prompt]
You are an expert education researcher coding essay feedback by the rhetorical strategy it uses to communicate to the student. The input is formatted with:
- EXCERPT: a window of essay text with brackets marking the highlighted portion
- COMMENT: the teacher feedback

Select every code that applies. Most comments will get exactly one label, but some legitimately combine more than one (for example, a comment that both praises and asks an open question, or one that both corrects an error and explains the underlying rule).

## Codebook:
\{
Label: 
Definition: 
Example: 
\}
\end{lstlisting}
\caption{System prompt for labeling essay feedback rhetorical strategy.}
\label{fig:feedback2-llm-judge-prompt}
\end{figure}

\clearpage
\section{Examples}
Tables~\ref{tab:vs-example-1}, \ref{tab:vs-example-3}, \ref{tab:sft-phrase-example}, \ref{tab:sft-phrase-example-2}, \ref{tab:math-vs-example-1}, \ref{tab:math-vs-example-2}, and \ref{tab:math-sft-example} compare examples of human-authored content and model outputs. We selected assignments and units with the largest diversity gaps and randomly sampled examples within each. Outputs exceeding 50 words were truncated with ellipses.
\begin{table*}[t]
\centering
\small
\begin{tabular}{@{}p{0.48\textwidth}|p{0.48\textwidth}@{}}
\toprule
\textbf{Human} & \textbf{Gemini-3.5-Flash + Verbalized sampling} \\
\midrule
\textbullet~Explain what the problem is that you're saying needs to be fixed. Is it kids getting inappropriate materials in books or the books being censored in general?\par\vspace{3pt}\textbullet~I appreciate that you are creating a solution: that's great, but make sure that you are also defining the issue and articulating what you believe about removing offensive materials before introducing your solution.\par\vspace{3pt}\textbullet~What are examples of things that might offend people and be in that room?\par\vspace{3pt}\textbullet~I appreciate that you're trying to signpost for your audience here: let's try to embed transition words and phrases rather than state them in sentences all on their own.\par\vspace{3pt}\textbullet~Always reread for grammar and mechanics\par\vspace{3pt}\textbullet~Okay, so hear me out: by including this line right after the idea that people would be old enough to handle the content, you're undercutting your argument because you're implying that actually age doesn't matter since what matters is whether someone gets offended easily.\par\vspace{3pt}\textbullet~Seems like this is the point in your writing that you're going to focus on your views on censorship\par\vspace{3pt}\textbullet~Rather than telling your audience that your idea is a good idea, you want to show them: what will your ideas enable?\par\vspace{3pt}\textbullet~I think your goal of persuasiveness is present throughout the text, but you can add more to your ideas & \textbullet~The contraction 'Thats' needs an apostrophe to become 'That's'.\par\vspace{3pt}\textbullet~Keep an eye on spelling! The correct spelling of this word is 'separate'.\par\vspace{3pt}\textbullet~Your spelling of 'seperate' needs a quick fix. Remember it is spelled 'separate'.\par\vspace{3pt}\textbullet~Spelling check: make sure to use an 'a' instead of an 'e' in 'separate'.\par\vspace{3pt}\textbullet~Correct 'Its' to 'It's' and adjust the apostrophe location in 'would'nt' to 'wouldn't'.\par\vspace{3pt}\textbullet~Please add an apostrophe to 'dont' to properly spell 'don't'. Also, consider using a more formal tone by avoiding 'I don't think'.\par\vspace{3pt}\textbullet~Review your spelling of 'Seperate'. The correct spelling is 'Separate'. Also, 'my views' is a bit vague for a concluding summary.\par\vspace{3pt}\textbullet~The correct spelling of 'Seperate' is 'Separate'. I encourage you to use more persuasive language than saying 'my views are all good ideas'.\par\vspace{3pt}\textbullet~Ensure correct word usage. 'Affective' has to do with emotions, whereas 'effective' denotes success at achieving a result. You want 'effective' here. \\
\bottomrule
\end{tabular}
\caption{Human vs.\ Gemini-3.5-Flash + verbalized sampling feedback on the same censorship essay.}
\label{tab:vs-example-1}
\end{table*}

\begin{table*}[t]
\centering
\small
\begin{tabular}{@{}p{0.48\textwidth}|p{0.48\textwidth}@{}}
\toprule
\textbf{Human} & \textbf{GPT-5.2 + Verbalized sampling} \\
\midrule
\textbullet~full of them vs. full of ``the'' - be mindful of proofreading as you look through your essay. Sometimes, spell check will miss errors when words are spelled correctly, even if they are misplaced/misused.\par\vspace{3pt}\textbullet~Perhaps you could give a specific example of distractions that impact driving quality - ex. look into research about texting while driving to make a stronger overall claim.\par\vspace{3pt}\textbullet~You should lead into the quote so that you integrate it into the language of your own paragraph. This will result in a smoother reading experience for your audience.\par\vspace{3pt}\textbullet~Be mindful of citations! Cite using the author's name, page number, etc. depending on the citation model you are using.\par\vspace{3pt}\textbullet~The above quotation does not speak about energy efficiency.\par\vspace{3pt}\textbullet~This offers stronger quotation integration in comparison to your first quotation. Good job!\par\vspace{3pt}\textbullet~The assignment instructions do not state it is necessary to include a counterpoint to your argument. What do you gain from the inclusion of this line?\par\vspace{3pt}\textbullet~Same comment as before about citations.\par\vspace{3pt}\textbullet~You have not yet dedicated enough time to the counterpoint you've raised! You should think about the ramifications of the question posed by the author of the original paper.\par\vspace{3pt}\textbullet~This is not a strong conclusion! Consider expanding the perspective you have on this issue by considering broader implications of this technology. & \textbullet~Add a specific example of a human driving flaw (from the article if possible) to make this point more concrete.\par\vspace{3pt}\textbullet~Add a transition that connects this idea to your previous point about human flaws to improve flow.\par\vspace{3pt}\textbullet~Consider using an example from the article (if it mentions texting, fatigue, or other distractions) to support this.\par\vspace{3pt}\textbullet~Try to include a statistic or example from the article to make this claim more persuasive.\par\vspace{3pt}\textbullet~Consider specifying a timeframe if the article provides one (ex: ``within the next decade'').\par\vspace{3pt}\textbullet~Provide a reason for this prediction (ex: testing, government regulations, insurance changes) or cite the article.\par\vspace{3pt}\textbullet~Clarify what ``another thing'' refers to-driving tasks, paying attention, reacting quickly-so the final point is specific. \\
\bottomrule
\end{tabular}
\caption{Human vs.\ GPT-5.2 + verbalized sampling feedback on the same driverless cars essay.}
\label{tab:vs-example-3}
\end{table*}

\begin{table*}[t]
\centering
\small
\begin{tabular}{@{}p{0.48\textwidth}|p{0.48\textwidth}@{}}
\toprule
\textbf{Human} & \textbf{Llama-3.1-8B + SFT} \\
\midrule
\textbullet~I like how you come out/on strong with your opinion, but your syntax creates confusion and detracts from the power of your claim.\par\vspace{3pt}\textbullet~Awk wording. Try: In paragraph four of the article, the author states, ``....''\par\vspace{3pt}\textbullet~Remember to unpack this a little bit with some analysis before you move onto your next point of support.\par\vspace{3pt}\textbullet~Never a good idea to end a paragraph on someone else's words. Remember the TEEAC or TIQA formula we studied in class.\par\vspace{3pt}\textbullet~A strong conclusion doesn't need this phrase (this is more middle school level). & \textbullet~I would combine these three sentences and rephrase the idea that there is no need for technology to detect students' emotional expressions.\par\vspace{3pt}\textbullet~Great start! \dag{}\par\vspace{3pt}\textbullet~I love that you are taking a stance on this issue! I love that you are citing textual evidence. I love that you are using transition words. What I don't love is that your evidence is not properly cited. Make sure you are citing the article correctly.\par\vspace{3pt}\textbullet~Good - maybe you can combine these lines to make one solid thesis?\par\vspace{3pt}\textbullet~This evidence also supports your argument. Great job! \\
\bottomrule
\end{tabular}
\caption{Human vs.\ Llama-3.1-8B + SFT feedback on the same Making Mona Lisa Smile essay.}
\label{tab:sft-phrase-example}
\end{table*}

\begin{table*}[t]
\centering
\small
\begin{tabular}{@{}p{0.48\textwidth}|p{0.48\textwidth}@{}}
\toprule
\textbf{Human} & \textbf{Llama-3.1-8B and Qwen-2.5-7B + SFT} \\
\midrule
\textbullet~I like that you have a celar position on the topic in the introduction paragraph, and that you have set up three clearly stated reasons.\par\vspace{3pt}\textbullet~You have done a good job of referring to a specific detail from the text to show that these cars can keep drivers alert and ready to take the wheel.\par\vspace{3pt}\textbullet~This reasons seems a lot like the previous paragraph's reason. Perhaps you can combine these two paragraphs into one. I like that you are using some of the text from the article as evidence and support.\par\vspace{3pt}\textbullet~Glad to see you are using a lot of text evidence. I would like to see you continue to work on integrating direct quotations into your own wrtiting. After a direct quotation, make sure you add your ideas and explanation.\par\vspace{3pt}\textbullet~As you come to the conclusion, make sure you restate your claim and summarize your main points. & \textit{Llama-3.1-8B}\par\vspace{2pt}\textbullet~Very nice job here! You are bringing in great points and organizing them in a clear and concise manner.\par\vspace{3pt}\textbullet~How does this support your claim that the car is not dangerous?\par\vspace{6pt}\textit{Qwen-2.5-7B}\par\vspace{2pt}\textbullet~This is a strong sub-claim!\par\vspace{3pt}\textbullet~Strong evidence from the text! How does this connect to your claim?\par\vspace{3pt}\textbullet~How does this sentence connect to your claim? \\
\bottomrule
\end{tabular}
\caption{Human vs.\ SFT feedback on the same driverless cars essay.}
\label{tab:sft-phrase-example-2}
\end{table*}

\begin{table*}[t]
\centering
\small
\begin{tabular}{@{}p{0.48\textwidth}|p{0.48\textwidth}@{}}
\toprule
\textbf{Human} & \textbf{GPT-5.2 + Verbalized sampling} \\
\midrule
\textbullet~Is the solution positive or negative? 2* x=6 -2* x=6.1 2.9 * x = -6.04 -2.473* x = -6.859\par\vspace{3pt}\textbullet~For each equation, write two more equations using the same numbers that express the same relationship in a different way. 3 + 2 = 5 7.1 + 3.4 = 10.5 15 - 8 = 7 3/2 + 9/5 = 33/10\par\vspace{3pt}\textbullet~In 2012, James Cameron descended to the bottom of Challenger Deep in the Marianas Trench; the deepest point in the ocean. The vessel he rode in was called DeepSea Challenger. Challenger Deep is 35,814 feet deep at its lowest point DeepSea Challenger’s descent was a change in depth of (-4) \ldots{}\par\vspace{3pt}\textbullet~Write a sentence to answer each question: How much warmer is 82 than 40? How much warmer is 82 than -40?\par\vspace{3pt}\textbullet~A submarine is only allowed to change its depth by rising toward the surface in 60-meter stages. It starts off at -340 meters. At what depth is it after: 1 stage 2 stages 4 stages How many stages will it take to return to the surface?\par\vspace{3pt}\textbullet~Some boats were traveling up and down a river. A satellite recorded the movements of several boats. A motor boat traveled -3.4 miles per hour for 0.75 hours. How far did it go? A tugboat traveled -1.5 miles in 0.3 hours. What was its velocity? What do you think that \ldots{}\par\vspace{3pt}\textbullet~Clare is cycling at a speed of 12 miles per hour. If she starts at a position chosen as zero, what will her position be after 45 minutes? Han is cycling at a speed of -8 miles per hour; if he starts at the same zero point, what will his \ldots{}\par\vspace{3pt}\textbullet~Find the value of each expression. 16.2 + -8.4 2/5 - 3/5 -9.2 + -7 -4 3/8 - (-1 1/4) & \textbullet~A scuba diver changes depth at a constant rate of -3 m/min for 7 min. What is the change in depth?\par\vspace{3pt}\textbullet~A cyclist’s change in position is -24 km after riding for 3 h at constant velocity. What was the cyclist’s velocity (km/h)?\par\vspace{3pt}\textbullet~Compute: (-1.5)(8)\par\vspace{3pt}\textbullet~Compute: (-7) + 12\par\vspace{3pt}\textbullet~Compute: (-3/4) + (5/6)\par\vspace{3pt}\textbullet~Compute: (-9)(-5)\par\vspace{3pt}\textbullet~Compute: (-2/3)(9/4)\par\vspace{3pt}\textbullet~Compute: (-7/10) + (-9/5) \\
\bottomrule
\end{tabular}
\caption{Human vs.\ GPT-5.2 + verbalized sampling problems for the same signed-number arithmetic unit.}
\label{tab:math-vs-example-1}
\end{table*}

\begin{table*}[t]
\centering
\small
\begin{tabular}{@{}p{0.48\textwidth}|p{0.48\textwidth}@{}}
\toprule
\textbf{Human} & \textbf{Gemini-3.5-Flash + Verbalized sampling} \\
\midrule
\textbullet~Solve each equation. 2(x-3) = 14 -5(x-1) = 40 12(x+10)=24 1/6(x+6)=11 5/7(x-9)=25\par\vspace{3pt}\textbullet~Elena walked 20 minutes more than Lin. Jada walked twice as long as Elena. Jada walked for 90 minutes. The equation 2(x+20)=90 describes this situation. Match each expression with the statement in the story with the expression it represents. A: x B: x+20 C: 2(x+20) D: 90 1: The number \ldots{}\par\vspace{3pt}\textbullet~Solve the inequality that represents each story. Then interpret what the solution means in the story. For every \$9 that Elena earns, she gives x dollars to charity. This happens 7 times this month. Elena wants to be sure she keeps at least \$42 from this month’s earnings. 7(9-x) >= \ldots{}\par\vspace{3pt}\textbullet~Andre wants to buy a backpack. The normal price of the backpack is \$40. He notices that a store that sells the backpack is having a 30\% off sale. What is the sale price of the backpack?\par\vspace{3pt}\textbullet~Match each story to an equation. A: A stack of nested paper cups is 8 inches tall. The first cup is 4 inches tall and each of the rest of the cups in the stack adds 1/4 inch to the height of the stack. B: A baker uses 4 cups \ldots{}\par\vspace{3pt}\textbullet~A certain shade of blue paint is made by mixing 1 1/2 quarts of blue paint with 5 quarts of white paint. If you need a total of 16.25 gallons of this shade of blue paint, how much of each color should you mix?\par\vspace{3pt}\textbullet~You know x is a number less than 4. Select all the inequalities that must be true. A: x<2 B: x+6<10 C: 5x<20 D: x-2>2 E: x<8\par\vspace{3pt}\textbullet~Match each sentence with the inequality that could represent the situation. A: Han got \$2 from Clare, but still has less than \$20. B: Mai spent \$2 and has less than \$20. C: If Tyler had twice the amount of money he has, he would have less than \$20. D: \ldots{} & \textbullet~Solve the inequality -3y - 4 >= 8 and state whether the endpoint on the number line is open or closed.\par\vspace{3pt}\textbullet~A tape diagram is partitioned into 4 equal parts labeled x, with an additional part labeled 12. The entire length of the tape diagram is 40. Write an equation that represents this tape diagram.\par\vspace{3pt}\textbullet~A number puzzle says: 'Multiply a number x by 3, add 12, and the result is 27.' Write an equation to represent this puzzle.\par\vspace{3pt}\textbullet~Find the value of m that makes the equation 1/4m - 5 = -2 true.\par\vspace{3pt}\textbullet~Write an equation that represents a balanced hanger diagram with 4 identical blocks of weight y and a block of weight 3 on one side, and a block of weight 15 on the other side.\par\vspace{3pt}\textbullet~Solve the inequality 2x + 5 < 13 and describe the solution on a number line.\par\vspace{3pt}\textbullet~A taxi service charges a flat fee of \$3.50 plus \$2.25 per mile. Write and solve an equation to find the number of miles, m, a passenger traveled if the total fare was \$26.00.\par\vspace{3pt}\textbullet~Solve the inequality -5(z + 2) < 15 and write the solution. \\
\bottomrule
\end{tabular}
\caption{Human vs.\ Gemini-3.5-Flash + verbalized sampling problems for the same equations and inequalities unit.}
\label{tab:math-vs-example-2}
\end{table*}

\begin{table*}[t]
\centering
\small
\begin{tabular}{@{}p{0.48\textwidth}|p{0.48\textwidth}@{}}
\toprule
\textbf{Human} & \textbf{Qwen-2.5-7B + SFT} \\
\midrule
\textbullet~Sixth-grade students were asked, “What grade are you in?” Explain why this is not a statistical question.\par\vspace{3pt}\textbullet~Zookeepers recorded the ages, weights, genders, and heights of the 10 pandas at their zoo. Write two statistical questions that could be answered using these data sets.\par\vspace{3pt}\textbullet~Mai and Priya each played 10 games of bowling and recorded the scores. Mai’s median score was 120, and her IQR was 5. Priya’s median score was 118, and her IQR was 15. Whose scores probably had less variability? Explain \ldots{}\par\vspace{3pt}\textbullet~Invent a data set with five numbers that has a mean of 10 and a median of 12.\par\vspace{3pt}\textbullet~Adult elephant seals generally weigh about 5,500 pounds. If you weighed 5 elephant seals, would you expect each seal to weigh exactly 5,500 pounds? Explain your reasoning. & \textbullet~A survey was conducted at your school to find out how many books each student read over the summer. Here are the results: \ldots{} Create a histogram to display this data. Then, describe the distribution of the number of books \ldots{}\par\vspace{3pt}\textbullet~A local park has been tracking the number of visitors each day for a week. Here are the numbers: \ldots{} 1. Make a histogram to represent this data. 2. Find the mean, median, and mode of the data. 3. Calculate \ldots{}\par\vspace{3pt}\textbullet~A local park has been recording the number of visitors each day for a week. Here are the numbers: \ldots{} Calculate the mean, median, and mode of the daily visitor counts.\par\vspace{3pt}\textbullet~A local pet store wants to know more about the types of pets their customers own. They decide to survey 50 customers to find out what kind of pets they have. Here are the results: \ldots{} Create a bar graph \ldots{}\par\vspace{3pt}\textbullet~A survey was conducted at a middle school to find out how many books each student read over the summer. Here are the results: \ldots{} Create a histogram to represent this data. Then, calculate the mean and median number of \ldots{} \\
\bottomrule
\end{tabular}
\caption{Human vs.\ Qwen-2.5-7B + SFT problems for the same statistics unit.}
\label{tab:math-sft-example}
\end{table*}

 % Reference the selected examples -- models even with diversity prompting repeat the request in the lesson narrative / prompt, whereas humans come up with different ways of asking the question (via choices), or with direct answer solilcitation etc. similar trends with essay feedback?
\end{document}